# Safety-Critical Human-Machine Shared Driving for Vehicle Collision Avoidance based on Hamilton-Jacobi reachability

Shiyue Zhao[1,2], Junzhi Zhang[1*], Rui Zhou[3], Neda Masoud[2], Jianxiong Li[1], Helai Huang[3], Shijie Zhao[1]

1 School of Vehicle and Mobility, Tsinghua University, Beijing 10084, China

2 Department of Civil and Environmental Engineering, University of Michigan, Ann Arbor, Michigan, 48105, USA.

3 School of Traffic and Transportation Engineering, Central South University, Changsha 410075, China

*Corresponding author: Junzhi Zhang (thu_znqzd@163.com)

*Current address: 2366 Erie Dr, Ann Arbor, Michigan, USA, 48105.

***Abstract***—Road safety continues to be a pressing global issue, with vehicle collisions imposing significant human, societal, and economic burdens. Human-machine shared collision avoidance in critical collision scenarios aims to aid drivers' accident avoidance through intervening only when necessary. Existing methods count on replanning collision-free trajectories and imposing human-machine tracking, which usually interrupts the driver's intent and increases the risk of conflict. This paper introduces a Reachability-Aware Reinforcement Learning (RL) framework for shared control, guided by Hamilton-Jacobi (HJ) reachability analysis. Machine intervention is activated only when the vehicle approaches the Collision Avoidance Reachable Set (CARS), which represents states where collision is unavoidable. First, we precompute the reachability distributions and the CARS by solving the Bellman equation using offline data. To reduce human-machine conflicts, we develop a driver model for sudden obstacles and propose an authority allocation strategy considering key collision avoidance features. Finally, we train a RL agent to reduce human-machine conflicts while enforcing the hard constraint of avoiding entry into the CARS. The proposed method was tested on a real vehicle platform. Results show that the controller intervenes effectively near CARS to prevent collisions while maintaining improved original driving task performance. Robustness analysis further supports its flexibility across different driver attributes.

## 1. Introduction

Road safety remains a critical global concern, as accidents and their consequences still pose a grave threat to individuals, communities, and economies. Statistics provided by transportation authority's display a considerable social and economic load—not only on human life and injury but also on whole financial and economic structures [1,2]. In safety-critical collision scenes where complex, timely, and precise operation is required, it is extremely difficult, even for well-experienced drivers, to make optimal decisions quickly [3]. Under emergency conditions, when immediate reactions are needed, cognitive overload and low situational awareness frequently inhibit a driver's ability to select the most appropriate evasion maneuvers to avoid collision, rendering it a significant challenge and a valuable research topic [4,5].

In recent years, both academia and industry have increasingly focused on human-machine shared control frameworks aiming to temporarily assist drivers in high-risk scenarios. Most current mainstream methods emphasize joint trajectory tracking, where upper-level planning provides collision-free trajectories and lower-level control ensures precise trajectory tracking by combining driver and automated actions. These methods implicitly assume that drivers willingly and smoothly accept new trajectories. However, in practice, drivers may instinctively or mistakenly resist system interventions, potentially causing excessive resistance or incorrect operations that introduce even greater collision risks [6,7].

Other approaches prioritize preserving the original intentions of drivers or the existing driving system, intervening only when necessary. Two prominent technical paths include model predictive control (MPC), which handles obstacles and risks as soft constraints but may prefer maintaining original trajectories under extreme conditions, potentially breaching safety limits. Alternatively, reinforcement learning (RL) methods often assess risks based on distance thresholds or potential field functions. However, intervention timing in these methods typically relies on empirical thresholds without strong theoretical foundations [8-12].

Considering these issues, this study seeks to introduce a theoretically sound, controllable intervention timing approach based on Hamilton-Jacobi (HJ) reachability analysis. Leveraging large-scale data and reinforcement learning (RL) techniques, this work aims to approximate collision avoidance reachable sets (CARS) and practically implement HJ-inspired constraints without incurring prohibitive computational costs. The key contributions are:

1) **An efficient data-driven method for solving collision avoidance Reachable Set (CARS),** leveraging data and RL to overcome the dimensionality challenges in HJ reachability analysis.
2) **A human-machine coordination mechanism**, integrating driver collision avoidance ability and intention to enable adaptive authority allocation and minimize human-machine conflicts.
3) **A reachability-aware RL framework**, embedding CARS-based constraints to ensure system safety while optimizing original task performance.
4) **Validation through real-world experiments,** highlighting the effectiveness, robustness, and adaptability of the proposed framework in high-risk driving situations.

## 2. Related Works

Traditional human-machine shared collision avoidance approaches, primarily based on path replanning and trajectory tracking, have demonstrated significant effectiveness in ensuring vehicle safety [8]. These approaches typically generate collision-avoidance trajectories and co-track with the driver through automatic intervention. For instance, Tsoi et al. [9] designed a haptic guidance system to support both lane-keeping and lane-changing tasks, providing continuous torque adjustments on the steering wheel for smoother lane transitions under obstacle avoidance scenarios. Wang et al. [10] proposed a shared steering control framework for high-speed emergency obstacle avoidance, utilizing fuzzy logic to dynamically allocate control authority between the driver and machine, combined with a nonlinear path tracking controller to enhance tracking precision. Lastly, Wu et al. [11] introduced a cooperative control strategy that integrates an improved collision avoidance path with a multi-constraint MPC-based yaw moment controller, ensuring safety and stability in high-speed collision avoidance situations. However, these trajectory-tracking-based strategies often fail to consider the driver’s real-time intentions, risking conflicts and introducing new collision risks [12].

To reduce human-machine conflicts, some studies have proposed human-centered shared driving methods [13-15]. Erlien et al. [14] introduced a steer-by-wire framework that leverages safe driving envelopes defined by vehicle handling limits and spatial constraints, employing a model predictive control (MPC) scheme with relaxation variables. Similarly, Song et al. [15] proposed a constrained MPC-based approach integrating vehicle safety and driver intentions, treating obstacle avoidance as soft constraints. However, soft constraint handling may compromise safety during critical human-machine interactions.

Several RL-based approaches generate collision avoidance maneuvers without explicitly designing dedicated obstacle-avoidance trajectories [16-20]. Yan et al. [17] developed a human-vehicle shared control framework leveraging imitation learning and RL for complex highway scenarios. Lv et al. [18] presented a safety-aware RL approach integrating a Frenet-based dynamic potential field model. Xu et al. [19] proposed a deep Q-network-based cooperative driving scheme for pedestrian collision avoidance. These methods primarily assess collision risks through distance-based or potential field methods, relying on empirical thresholds for intervention timing and lacking stringent theoretical guarantees [20].

Despite significant advancements, current methods still face inherent limitations. Trajectory-centric strategies risk imposing predetermined paths, causing driver conflicts [12]. Approaches that soften obstacle constraints can compromise safety, and RL methods lack theoretical justification for intervention timing [20]. Hamilton-Jacobi (HJ) reachability analysis provides a rigorous foundation for strict safety constraints but faces computational challenges in high-dimensional vehicle dynamics, known as the "curse of dimensionality" [21-23].

## 3. Framework Overview and Theoretical Foundations

To address the challenges of collision avoidance in human-machine shared control, this research integrates driver behavior modeling, reinforcement learning, and reachability analysis into a unified framework. The framework is designed to ensure safety while minimizing machine interventions, effectively balancing driver intent with automated control actions. At the core of this framework lies Hamilton-Jacobi (HJ) reachability, which reasonably quantifies the ability of a vehicle to escape obstacles in its environment from a given state. Based on this analysis, the Collision Avoidance Reachable Set (CARS) is defined, representing states from which future collisions are inevitable.

By embedding reachability analysis into reinforcement learning, the framework enforces safety as a hard constraint, guiding the learning process to avoid unsafe states while optimizing control strategies to minimize machine intervention. This chapter provides an overview of the framework's architecture, detailing the key modules and their interactions, and introduces the theoretical foundation of reachability analysis and its role in shared control systems.

### *3.1 Framework Overview*

Inspired by the concept of reachability in safe reinforcement learning, we propose a novel method to define the necessity of machine intervention based on obstacle reachability in emergency collision avoidance scenarios. Specifically, in critical situations involving obstacles, if no optimal control policy exists to prevent the vehicle from colliding, the current state of the vehicle is deemed to belong to the Collision Avoidance Reachable Set (CARS), formally solved in Section III. Naturally, the proximity of a vehicle's state to the CARS boundary serves as a critical criterion for determining the necessity of machine intervention. For instance, when the vehicle is far from the CARS boundary, the driver is likely able to avoid the collision autonomously based on their own intent. From the perspective of minimizing human-machine conflicts, there is no need to apply machine intervention or alter the original trajectory in such scenarios.

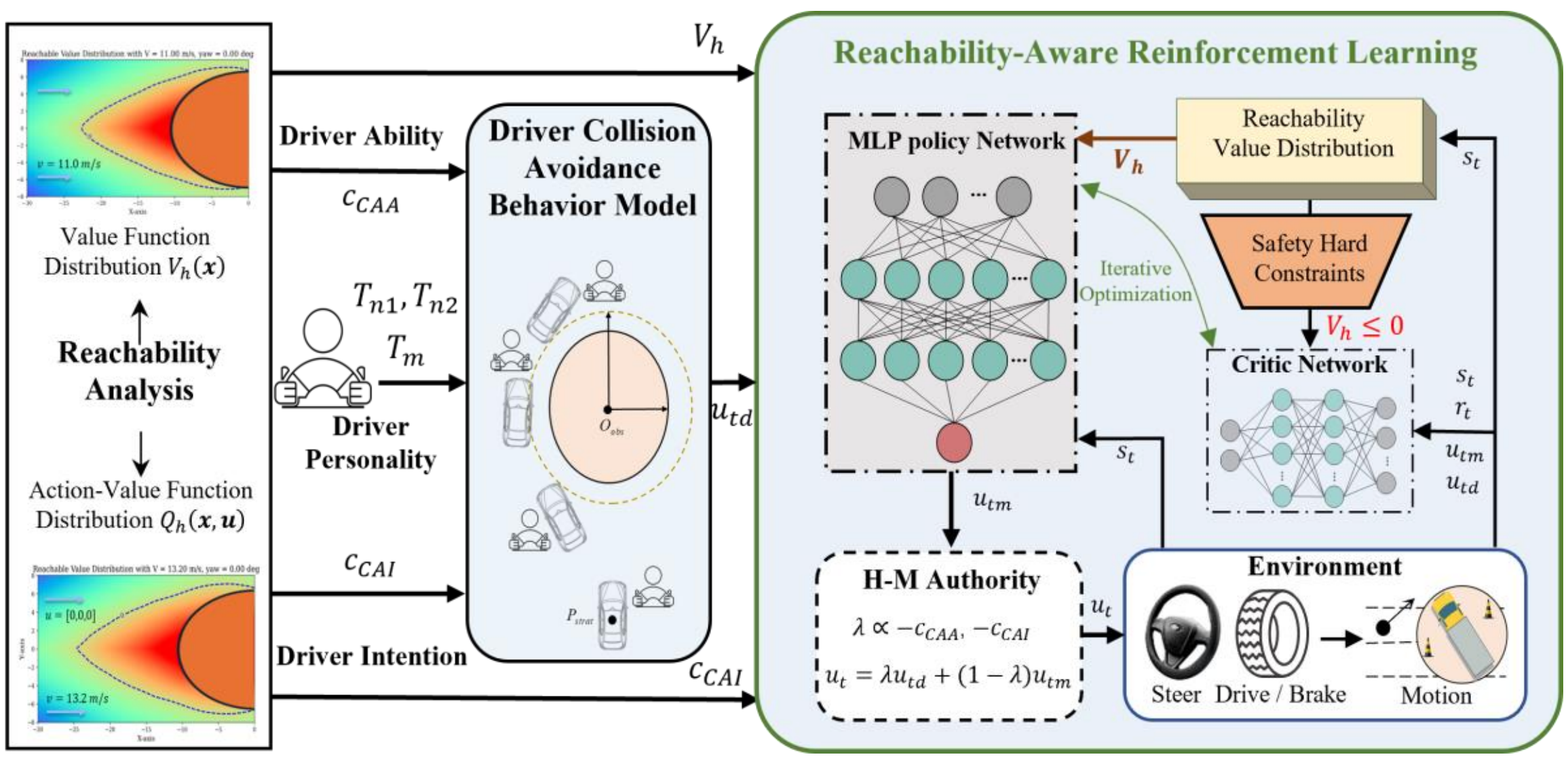

**Figure 1**: Reachability-Aware Reinforcement Learning Framework.

This paper introduces the Reachability-Aware RL Framework (Figure 1), designed to reduce human-machine conflicts while enforcing the hard constraint of avoiding entry into the CARS. The framework consists of two main phases:

**Offline Learning:** inspired by methods in [20,24], we precompute reachability distributions and CARS using collected data and offline RL methods, overcoming the curse of dimensionality. Additionally, a driver behavior model is built to provide driver actions for online learning.

**Online Learning:** Illustrated on the left side of Figure 1, the reachability value of the current vehicle state is used to help the RL agent assess the state's escape potential relative to obstacles. The CARS boundary serves as a hard safety constraint to ensure the agent avoids unsafe states [20]. A human-machine authority allocation mechanism dynamically balances driver intent with state reachability. The policy network, which generates machine actions, is updated using the Actor-Critic algorithm.

During deployment, the driver intent recognition module analyzes driver actions, combining them with vehicle state and reachability values as inputs to the policy network. Final control commands are determined by the authority allocation mechanism.

### *3.2 Hamilton-Jacobi Reachability: Theoretical Foundations*

Hamilton-Jacobi (HJ) reachability offers a framework for analyzing system safety by identifying states from which collisions or other unsafe conditions are unavoidable. A key element of this framework is the function $h(x)$, which represents the safety constraint for a given state $x$. Specifically, $h(x) > 0$ indicates that the state is unsafe, while $h(x) < 0$ signifies that the state is within the safe region. In this study, h(x) is used to define the states where collisions occur, making it central to determining the safety boundaries in dynamic systems. HJ reachability computes value functions through solving a Hamilton-Jacobi partial differential equation (HJ PDE), quantifying the system's ability to avoid unsafe states. These value functions are crucial for defining and enforcing safety boundaries, particularly in collision avoidance scenarios.

Definition 1: Optimal Reachable Value Function and Action-Value Function

In collision avoidance scenarios, a key challenge is to evaluate whether a vehicle starting from a certain state can avoid entering unsafe regions such as collisions under any admissible control strategy. To quantify this, we define two value functions based on the concept of Hamilton-Jacobi (HJ) reachability analysis. These functions quantify the worst-case safety margin along possible collision avoidance trajectories under optimal control.

Let $h(x)$ denote the safety constraint function, where $h(x) \geq 0$ denotes unsafe states (e.g., collision states) and $h(x) < 0$ represents safe states. Let $x_\tau$ denote the discrete state of the system at time $\tau$ under control inputs. $\mathcal{U}$ is the set of all admissible control inputs, and the index $\tau \in N$ refers to the discrete time steps.

HJ Reachability defines the optimal reachable state-value function $V_h(x)$ as:

$$V_h(x) = \min_{u(\cdot)\in\mathcal{U}}[\max_{\tau\in N} h(x_\tau)]\ ,\ x(0) = x \tag{1}$$

which computes, for a given initial state $x$, the smallest possible upper bound of constraint violation along all future trajectories. The outer minimization seeks the control sequence that best mitigates the worst-case safety risk, while the inner maximization identifies the moment along the trajectory at which the system is closest to becoming unsafe.

To evaluate the safety consequences of a specific immediate control action, HJ Reachability further define the optimal reachable action-value function as:

$$Q_h(x,u) = \min_{u(\cdot)\in\mathcal{U}}[\max_{\tau\in N} h(x_\tau)]\,,\ x_0 = x, u_0 = u \quad (2)$$

which assesses the safety margin starting from state $x$, when the first control input is fixed as $u$, followed by an optimal policy thereafter. This function enables action-level safety evaluation and can guide the design of real-time safe action filters or backup controllers in uncertain environments.

Here, These value functions, computed on a discrete grid, possess the following properties:

- $V_h(\boldsymbol{x}) < 0$: Indicates that a policy exists to keep the system state safe (i.e., avoid unsafe states) starting from $x$.
- $V_h(\boldsymbol{x}) \geq 0$: Indicates that no policy can prevent the system from entering unsafe states starting from $x$.

**Definition 2:** Collision Avoidance Reachable Set - CARS

Similar to the definition of the **Backward Reachable Set** in Hamilton-Jacobi reachability analysis [22,25], the Collision Avoidance Reachable Set (CARS) represents the set of states from which it is impossible to avoid a collision with obstacles, regardless of the control input applied. It is defined as:

$$CARS = \boldsymbol{x} \in \mathbb{R}^n \mid V_h(\boldsymbol{x}) > 0 \quad (3)$$

In this context:

- $x \in CARS$: Indicates that the system is in an unsafe region, and a collision is inevitable under any control policy.
- $x \notin CARS$: Indicates that there exists a control policy that can steer the system away from unsafe states.

States closer to the CARS boundary signify higher risk and require more immediate and stringent control actions.

## 4. Authority Allocation Strategy Based on HJ Reachability

This section introduces an authority allocation strategy for human-machine shared control based on Hamilton-Jacobi reachability. The CARS is computed to define safety boundaries, while driver intention and ability models are proposed to evaluate the driver's decision-making and control capability. The driver's ability is directly linked to the system's reachability, reflecting the state's potential to avoid collisions.

### *4.1 Reachability Analysis and CARS Computation*

The computation of the reachability value functions $V_h(x)$ and $Q_h(x, a)$, as defined in Definition 1, often relies on dynamic programming approaches. However, such methods require precise vehicle dynamic models, which can introduce inaccuracies in real-world scenarios. Moreover, using a sixth-order vehicle dynamics model in dynamic programming leads to severe dimensionality issues, commonly referred to as the curse of dimensionality. To address this, Fisac et al. [26] proposed a novel approach that connects HJ reachability with reinforcement learning by introducing a discount factor $\gamma \rightarrow 1$ into $Q_h$. Specifically, they reformulate the safety problem as a discounted minimum-payoff optimal control task, yielding a Safety Bellman Equation that incorporates time-discounting to enable the use of temporal difference learning techniques. Moreover, Fisac et al. [26] prove that $lim_{\gamma \rightarrow 1} Q_{h,\gamma} \rightarrow Q_h$ provided $h$ is bounded, , so the discounted solution approximates the classical undiscounted HJ reachability value with arbitrary accuracy by annealing $\gamma$ toward 1 during training. This formulation yields a reachability Bellman operator $\mathcal{B}^*$ [26], as shown in Equation 4-5.

$$\mathcal{B}^* Q_{h,\gamma}(\boldsymbol{x}, \boldsymbol{u}) \coloneqq (1-\gamma)h(\boldsymbol{x}) + \gamma \max\{h(\boldsymbol{x}), V_{h,\gamma}^*(\boldsymbol{x}')\} \tag{4}$$

$$V_{h,\gamma}^*(\boldsymbol{x}') = \min_{a'} Q_{h,\gamma}(\boldsymbol{x}', \boldsymbol{u}') \tag{5}$$

Here, $x'$ represents the successor state reached from state $x$ after applying an action $u$.

To substantiate the theoretical validity of this conversion from HJ reachability to a Q-function framework with $\gamma \rightarrow 1$, Fisac et al. [26] rigorously prove that the discounted Safety Bellman operator $\mathcal{B}^*$ induces a contraction mapping under the supremum norm for any $\gamma \in [0,1)$, ensuring the existence of a unique fixed-point solution and enabling convergence of iterative methods like value iteration. Furthermore, they establish the convergence of the resulting Safety Q-learning algorithm (Theorem 2 in [26]) in finite-state finite-action Markov decision processes (MDPs), with probability 1, under standard assumptions such as infinite exploration, learning rates $\alpha_k$ satisfying $\sum \alpha_k = \infty$ and $\sum \alpha_k^2 < \infty$, and partial randomization in episode initialization and policy selection. This convergence holds to the optimal discounted state-action safety value function, and by annealing $\gamma$ asymptotically to 1, the method recovers the undiscounted HJ safety value function with resolution completeness. Neural network approximators in high dimensions (e.g., up to 18D in [26]) lack absolute convergence guarantees due to unproven RL training, potentially causing errors or local optima. However, empirical validation in [20,24,26] shows high accuracy (e.g., misclassification rates below $10^{-3}$ and conservative safety approximations via forward simulation of the learned policy, mitigating risks in deployment.

To address discrepancies in the simulator model, we adopt an offline learning framework like [20], utilizing real-world data collected from the experimental platform described in Section V.B. This dataset includes multi-strategy collision avoidance data from four drivers, encompassing over 300 instances with diverse initial conditions, such as varying positions, speeds, strategies, and yaw angles. After applying data augmentation techniques, the dataset size increases to 31 million samples. During data collection, the vehicle's ESC system operated normally, with non-collision episodes accounting for 65% of the samples and collision episodes making up the 35%. In the context of collision avoidance, the state $x$ is defined as:

$$\boldsymbol{x} = \left[X, Y, \varphi, v_x, v_y, r\right] \tag{6}$$

where $X$ and $Y$ are the vehicle's global position coordinates, $\varphi$ is the yaw angle, $v_x$ and $v_y$ are the longitudinal and lateral velocities, respectively, and $r$ represents the yaw rate.

The optimal reachability value functions, $V_{h,\gamma}$ and $Q_{h,\gamma}$, are then learned by minimizing the following loss functions:

$$\mathcal{L}_{V_{h,\gamma}} = \mathbb{E}_{(\boldsymbol{x},\boldsymbol{u})\sim\mathcal{D}}\left[L_{rev}^{\upsilon}\left(Q_{h,\gamma}(\boldsymbol{x},\boldsymbol{u}) - V_{h,\gamma}(\boldsymbol{x})\right)\right] \tag{7}$$

$$\mathcal{L}_{Q_{h,\gamma}} = \mathbb{E}_{(\boldsymbol{x},\boldsymbol{u},\boldsymbol{x}')\sim\mathcal{D}}\left[\Big((1-\gamma)h(\boldsymbol{x}) + \gamma max\{h(x), V_{h,\gamma}(\boldsymbol{x}')\} - Q_{h,\gamma}(\boldsymbol{x},\boldsymbol{u})\Big)^2\right] \tag{8}$$

Here, $L_{\text{rev}}^{\upsilon}(\varepsilon) = |\upsilon - \mathbb{I}(\varepsilon > 0)|\varepsilon^2$ is an asymmetric loss function designed, giving greater emphasis to smaller values. By iteratively minimizing these loss functions, the approximate optimal reachability value functions are obtained, satisfying $lim_{\gamma\to 1}V_{h,\gamma} \to V_h$.

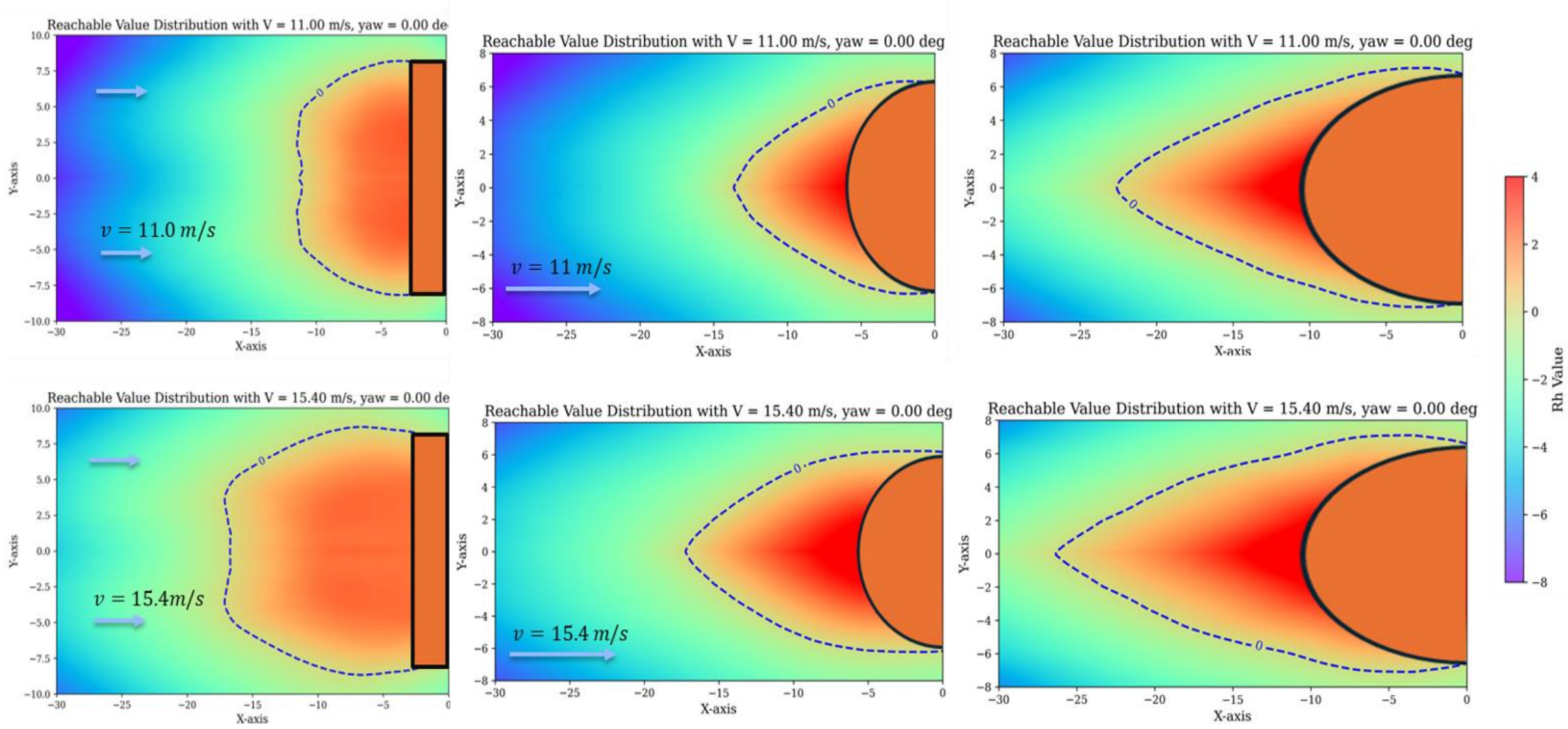

**Figure 2:** Reachability value distributions $V_h(x)$ and CARS boundaries (blue dashed lines) for (a) T-shaped, (b) circular, and (c) elliptical obstacles at varying velocities.

With the trained $V_h(\boldsymbol{x})$, we can efficiently evaluate the reachability of each state $x$ and subsequently approximate the CARS using Definition 2. Using the proposed method, we computed the reachability value function $V_h(\boldsymbol{x})$ and the approximate CARS for specific velocities and yaw angles across various obstacle geometries.

In Figure 2, the blue dashed lines represent the computed Collision Avoidance Reachable Set (CARS) boundaries under different obstacle geometries and vehicle speeds. The color map visualizes the reachability value distribution $V_h(x)$, where lower values (more negative) indicate states with greater collision avoidance potential, and values closer to zero imply higher risk of collision.

For the T-shaped obstacle (Figure 2a), the CARS boundary exhibits a near-planar profile at lower speeds (11.0 m/s), with risk concentrated around the vertical bar and junction area of the "T". As the speed increases (15.4 m/s), the reachable region contracts and the CARS boundary expands backward along the longitudinal axis, indicating that higher velocities reduce the available reaction space and lead to earlier intervention requirements.

For the T-shaped obstacle (Figure 2a), the CARS boundary exhibits a near-planar profile at lower speeds (11.0 m/s), with risk concentrated around the vertical bar and junction area of the "T". As the speed increases (15.4 m/s), the reachable region contracts and the CARS boundary expands backward along the longitudinal axis, indicating that higher velocities reduce the available reaction space and lead to earlier intervention requirements.

For the circular obstacle (Figure 2b), the reachability field is radially symmetric, and the CARS boundary maintains a concentric shape centered around the obstacle. This reflects the uniform risk distribution imposed by a symmetric geometry. Increasing velocity again shifts the CARS boundary outward, shrinking the safe region.

These results demonstrate that the proposed framework can adapt to different obstacle shapes and speeds, effectively capturing how both factors influence the reachability distribution and CARS boundaries. The consistent variation across scenarios shows that the learned reachability model is suitable for real-time safety assessment.

### *4.2 Collision Avoidance Ability and Intention Modeling*

This section evaluates the driver's collision avoidance ability and intention, which are key factors for adaptive human-machine shared control.

In human-machine shared control, driver ability in trajectory-following tasks is often quantified by tracking error. Similarly, we use Hamilton-Jacobi (HJ) reachability to evaluate collision avoidance ability (*CAA*), with the current state's reachability value indicating its risk level. Lower values correspond to stronger avoidance ability, providing a theoretically grounded measure of the driver's capability to mitigate long-term collision risks. The calculation of Collision Avoidance Ability (CAA) is defined as:

$$CAA = \frac{1}{1 + \alpha_{CAA} \cdot (V_h(\boldsymbol{x}) + C_{CAA})} \tag{9}$$

where $V_h(x)$ represents the reachability value of the current state, $C_{CAA}$ is a constant offset, and $\alpha_{CAA}$ is a scaling parameter that adjusts the sensitivity of the ability measure.

The driver's collision avoidance intention (CAI) is directly related to their driving actions, reflecting the extent to which their inputs consider collision avoidance. In this study, the driver's actions are represented as $\boldsymbol{u_d} = [\delta_f, T_d]$, where $\delta_f$ denotes the front-wheel steering angle, and $T_d$ represent the motor drive/brake torque output by driver operation.

Using the pre-trained reachability distribution networks, the state-action value function $Q_h(\boldsymbol{x}, \boldsymbol{u_d})$ evaluates the cost (i.e., reachability) of the driver's action $\boldsymbol{u_d}$ in the current state $\boldsymbol{x}$. Meanwhile, the state value function $V_h(\boldsymbol{x})$ estimates the minimal cost for the current state.

By taking the ratio of these two values, the deviation of the driver's action from the optimal collision avoidance action can be quantified. Since $V_h(\boldsymbol{x}) < 0$ is treated as a hard constraint in this study, with smaller $V_h(\boldsymbol{x})$ values (larger absolute magnitudes) indicating safer states, using $V_h(\boldsymbol{x})$ as the denominator is appropriate. The CAI is defined as:

$$CAI = \frac{Q_h(\boldsymbol{x}, \boldsymbol{u_d})}{\min_u Q_{h,\gamma}(\boldsymbol{x}', \boldsymbol{u}) Q_h} = \frac{Q_h(\boldsymbol{x}, \boldsymbol{u_d})}{V_h(\boldsymbol{x})}, 0 \leq CAI \leq 1 \tag{10}$$

The range of $CAI$ is constrained to $[0,1]$ to handle rare cases where $Q_h(\boldsymbol{x}, \boldsymbol{u_d}) > 0$, setting $CAI = 0$ under such conditions. Additionally, it is observed that when the vehicle is far from the obstacle, the large absolute value of $V_h(\boldsymbol{x})$ reduces $CAI$'s sensitivity to $\boldsymbol{u_d}$. Conversely, as the vehicle approaches the obstacle, the absolute value of $V_h(\boldsymbol{x})$ decreases, increasing $CAI$'s sensitivity to the driver's actions. This property enhances the design of the subsequent human-machine authority allocation mechanism.

#### *4.3 Human-Machine Authority Allocation*

In this study, we emphasize the importance of designing an authority allocation mechanism **decoupled** from the subsequent RL process, as this significantly enhances the system's interpretability. To achieve this, we propose a human-machine authority allocation mechanism that considers both collision avoidance ability (CAA) and collision avoidance intention (CAI). The machine intervention weight $\gamma$ is computed as:

$$\gamma = \max\big(\gamma_{\min}, (1 - s_{CAI})(1 - s_{CAA})\big) \tag{11}$$

where:

$$s_{CAI} = \frac{1}{1 + \exp\big(-k_{CAI1}(CAI - k_{CAI2})\big)}$$

$$s_{CAA} = \frac{1}{1 + \exp\big(-k_{CAA1}(CAA - k_{CAA2})\big)}$$

Parameters $k_{CAI1}$ and $k_{CAA1}$ control the slope of the sigmoid mappings $s_{CAI}$ and $s_{CAA}$. Large values make the curve almost step-like, enabling a rapid hand-over to the automation in time-critical situations; smaller values flatten the curve, allowing a smoother, driver-centred transition when urgency is low.

The companion parameters $k_{CAI2}$ and $k_{CAA2}$ define the **mid-point** of each sigmoid—the CAI or CAA value at which $s_{CAI}$ and $s_{CAA}$ reach 0.5. These thresholds are crucial for determining when the system considers the driver's input sufficiently reliable, adjusting the sigmoid curve left or right.

Together, these four hyperparameters enable fine-tuning of the system's response to driver behavior and collision risks, ensuring that the mechanism remains adaptive to varying scenarios—higher $k_1$ values amplify gamma's sensitivity (e.g., rapid decrease when CAI/CAA exceeds thresholds), while adjusting $k_2$ influences the baseline intervention level. The minimum machine intervention weight $\gamma_{\min}$ ensures that the machine remains actively involved in shared control, even in high-risk situations.

These parameters are determined before training the reinforcement learning network to generate the intervention action $\boldsymbol{u_m}$. During actual deployment, any adjustments to these parameters would require retraining the model to ensure consistency between the deployment and training environments.

In this paper, following the principle of grid search optimization, the same set of parameter values is used across all methods: $k_{CAI1} = 6$, $k_{CAA1} = 3$, $k_{CAI2} = 0.8$, $k_{CAI2} = 0.6$. Figure 3 shows how the machine intervention weight $\gamma$ increases with higher Collision Avoidance Intention ($CAI$)or Ability ($CAA$) with selected parameters, indicating reduced machine control when the driver is more capable or intentful. The top plot illustrates that γ is more sensitive to CAI in the high-intention region due to the shape of the sigmoid function. This allows the controller to quickly reduce intervention once the driver's intention becomes clear. The bottom plot shows a similar decreasing trend for increasing CAA, but with a slightly more linear profile, suggesting more gradual adaptation to changes in CAA.

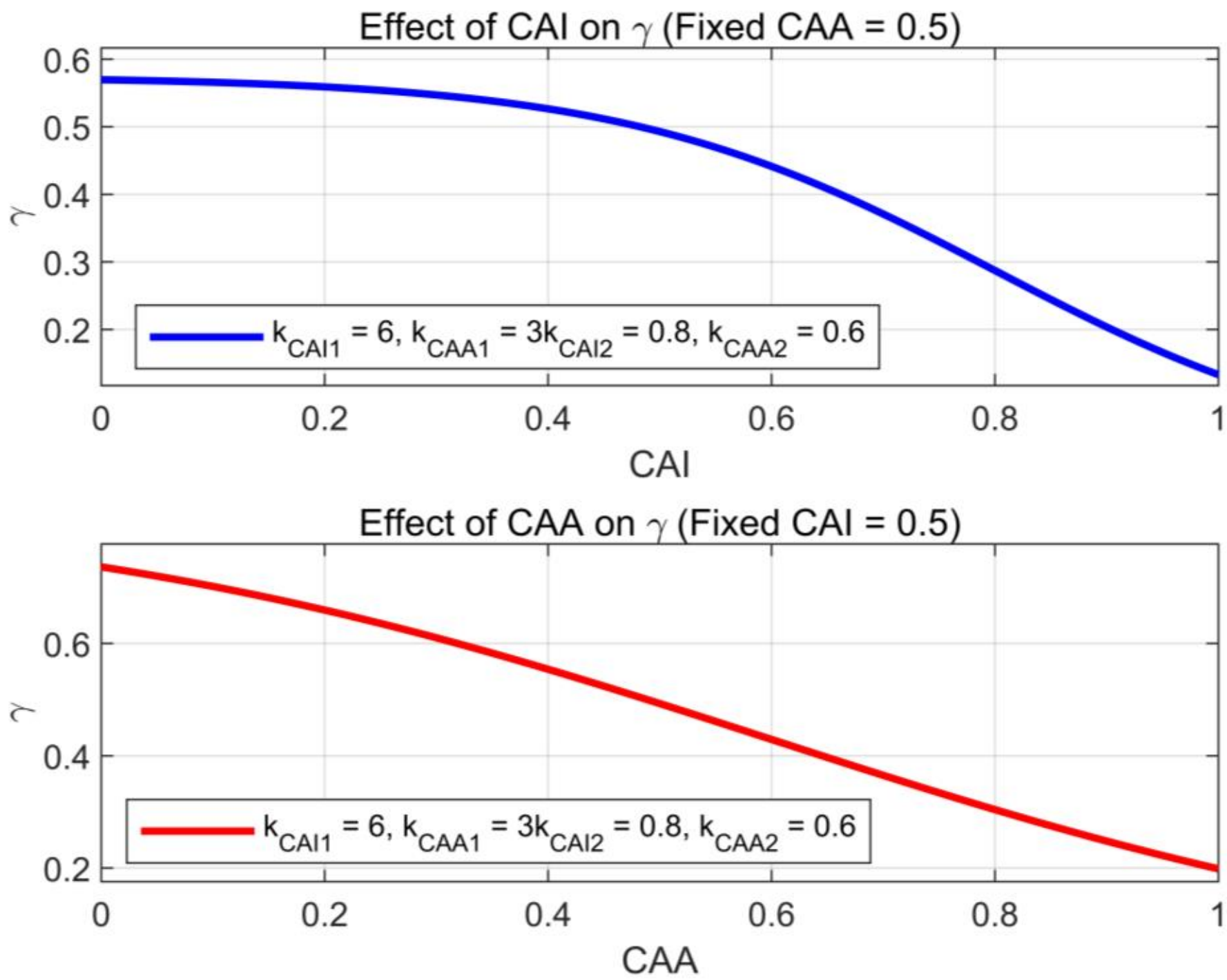

**Figure 3:** Variation of the machine intervention weight ($\gamma$) as functions of CAI and CAA.

The final shared control input $\boldsymbol{u_f}$ is then computed as:

$$\boldsymbol{u_f} = \gamma \cdot \boldsymbol{u_m} + (1 - \gamma) \cdot \boldsymbol{u_d} \tag{12}$$

where $\boldsymbol{u_m}$ is the machine-generated action derived through reinforcement learning in Section IV.

## 5. Reachability-Aware RL for Machine Action Generation

This section describes the design and training of a reachability-aware RL agent for generating safe and adaptive machine actions $\boldsymbol{u_m}$. Building on the previously defined reachability-based safety metrics $V_h(\boldsymbol{x})$ and $Q_h(\boldsymbol{x}, \boldsymbol{u})$, we embed these into the RL training process to maintain distance from the Collision Avoidance Reachable Set (CARS) and reduce human-machine conflicts.

This approach is based on a driver model for human-like behavior and a vehicle dynamics model for realistic responses. The obstacles are assumed to be elliptical, which simplifies analysis.

### *5.1 Driver Model for Human Action Simulation*

The driver model is essential for simulating realistic human actions during reinforcement learning training, allowing the RL agent to adapt to human-like behavior in collision avoidance scenarios. We propose a model based on driver preview behavior, focusing on how drivers anticipate and respond to obstacles.

In most collision avoidance scenarios, obstacle shapes can be approximated as ellipses, providing an accurate representation of the spatial relationship between the vehicle and obstacles. To simplify the subsequent analysis, we define the obstacle region (i.e., the unsafe state region) as the set $T$, where the elliptical obstacle is implicitly described by the function $h(x)$ as follows:

$$\boldsymbol{T} = \left\{ \boldsymbol{x} \,\middle|\, \boldsymbol{h}(\boldsymbol{x}) = \frac{(\boldsymbol{X}-\boldsymbol{X_0})^2}{\boldsymbol{a}^2} + \frac{(\boldsymbol{Y}-\boldsymbol{Y_0})^2}{\boldsymbol{b}^2} - \mathbf{1} < \mathbf{0} \right\} \qquad (\mathbf{13})$$

Here, $(X_0, Y_0)$ denotes the center coordinates of the ellipse, and $a$ and $b$ are the lengths of the semi-major and semi-minor axes, respectively.

Based on the collision avoidance data described in Section III.A, we analyzed the relationship between driver actions, collision intentions, and obstacle characteristics. Successful collision avoidance typically consists of two phases: (1) the obstacle avoidance phase, where the driver maneuvers around the obstacle, and (2) the recovery phase, where the driver realigns with the original trajectory after avoiding the obstacle. For unsuccessful attempts, only the first phase is observed as the vehicle fails to clear the obstacle.

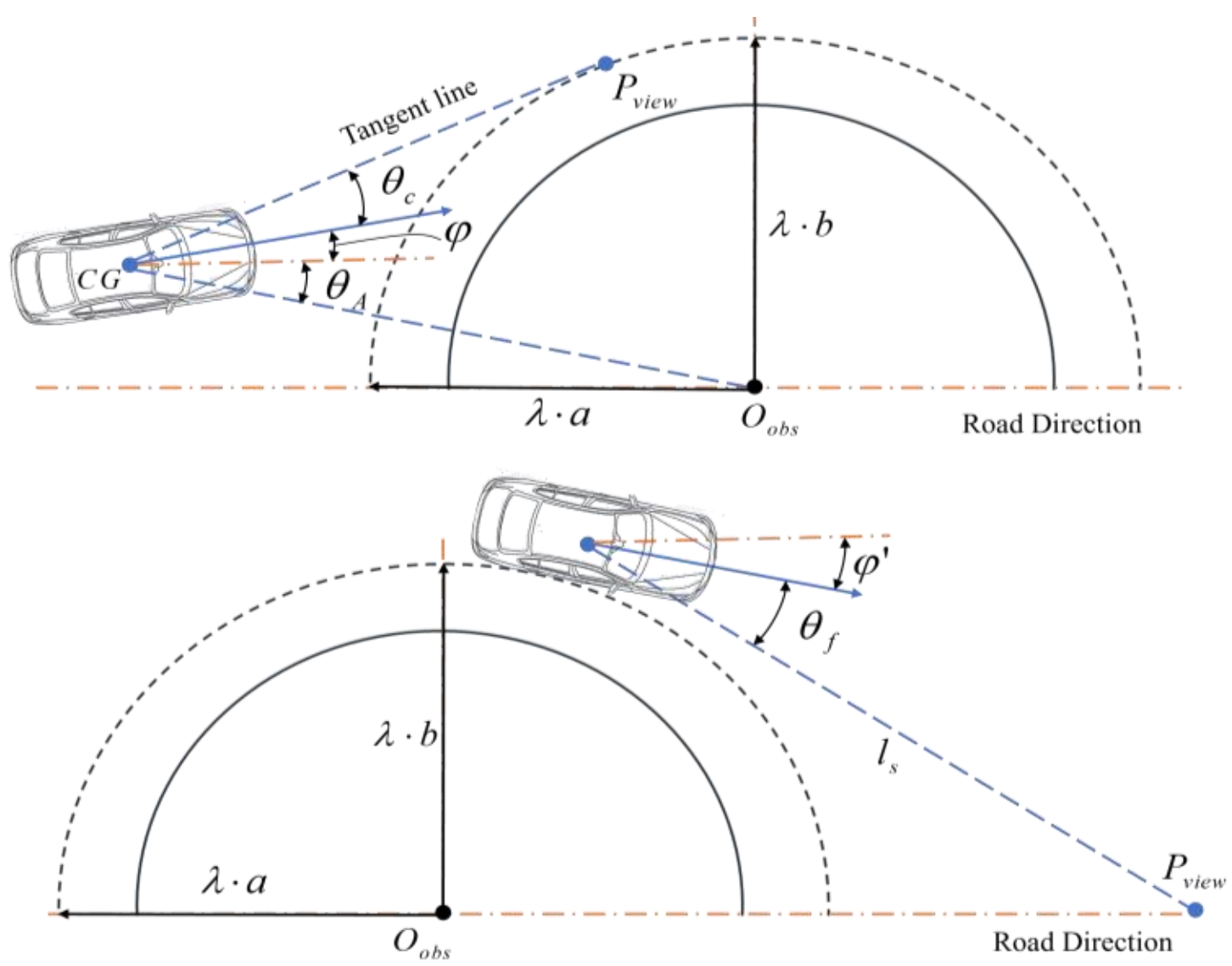

**Figure 4:** Illustration of Preview Angles during Obstacle Avoidance and Recovery Phases.

In the first phase, the driver's steering behavior can be represented as maneuvering around an elliptical envelope slightly larger than the obstacles, as shown in the upper half of Figure 4. The driver's preview direction aligns with the tangent of this elliptical boundary, guiding their steering input [27]. The preview angle $\theta_c$ is given by:

$$\theta_c = arctan\left(\frac{Y_{trg} - Y}{X_{trg} - X}\right) - \varphi + \mathcal{N}(0, \sigma^2) \tag{14}$$

We designed gaussian noise $\mathcal{N}$ to approximate the driver's realistic response. Here, and $(X_{trg}, Y_{trg})$ denotes the tangential point on the ellipse, which lies on the elliptical boundary defined by:

$$\frac{(X_{trg} - X_0)^2}{(\lambda \cdot a)^2} + \frac{(Y_{trg} - Y_0)^2}{(\lambda \cdot b)^2} - 1 = 0 \tag{15}$$

where $\lambda$ is a scaling factor that adjusts the ellipse size based on HJ reachability $V_h(\boldsymbol{x})$ and collision avoidance intention $CAI$, which is defined as:

$$\lambda = k_\lambda \cdot \frac{CAI}{|V_h(\boldsymbol{x})|} \tag{16}$$

Here, the parameter $k_\lambda$ controls the sensitivity of the ellipse size to $V_h(\boldsymbol{x})$ and $CAI$.

In the recovery phase, the driver transitions from avoiding the obstacle to returning to the original trajectory. Following the one preview point [6, 28], the driver focuses on a target point along the trajectory to compute their steering angle. As shown in the lower half of Figure 4, the recovery preview angle $\theta_f$ is calculated as:

$$\theta_f = arctan\left(\frac{\Delta Y}{l_s}\right) + e_\varphi \tag{17}$$

where $\Delta Y$ is the lateral distance from the vehicle to the target trajectory, $l_s$ is the fixed preview distance, and $e_\varphi$ is the heading error representing the angular deviation between the vehicle's orientation and the trajectory.

The conversion of the driver's visual preview angle $\theta$ to the front-wheel steering angle $\delta_d$ involves two primary stages: cognitive delay and neuromuscular response. The cognitive delay $T_m$ introduces a lag in processing the preview angle, modeled as:

$$\theta_p(t) = \begin{cases} \theta_c(t - T_m) & in\ collision\ phase \\ \theta_f(t - T_m) & in\ recovery\ phase \end{cases} \tag{18}$$

where $\theta_p(t)$ is the delayed angle passed to the neuromuscular system. The neuromuscular response is described by a first-order transfer function [29]:

$$G(s) = \frac{T_{n1}s + 1}{T_{n2}s + 1} \tag{19}$$

Here, the parameter $T_{n1}$ and $T_{n2}$ represent aspects of the driver's neuromuscular response, influencing the system's dynamics through the zero and pole. Subsequently, the steering angle $\delta_d$ is computed by:

$$\delta_d(s) = G(s) \cdot \Theta_p(s) \tag{20}$$

where $\delta_d(s)$ is the Laplace transform of the front-wheel steering angle, and $\Theta_p(s)$ is the Laplace transform of $\theta_p(t)$.

The driver's longitudinal control behavior is modeled by tracking a target speed:

$$v_{td} = \begin{cases} \alpha_{vd} \cdot v_{ori} \cdot CAI & if\ V_h(\boldsymbol{x}) \leq -3 \\ 0 & if\ V_h(\boldsymbol{x}) > -3 \end{cases} \tag{21}$$

where $v_{td}$ is the target speed, $\alpha_{vd}$ is a scaling factor, $v_{\text{ori}}$ is the original speed, and $CAI$ represents the driver's collision avoidance intention. A PID controller simulates the driver's rear axle torque inputs $T_d$ to track $v_{td}$, effectively modeling longitudinal behavior in collision avoidance.

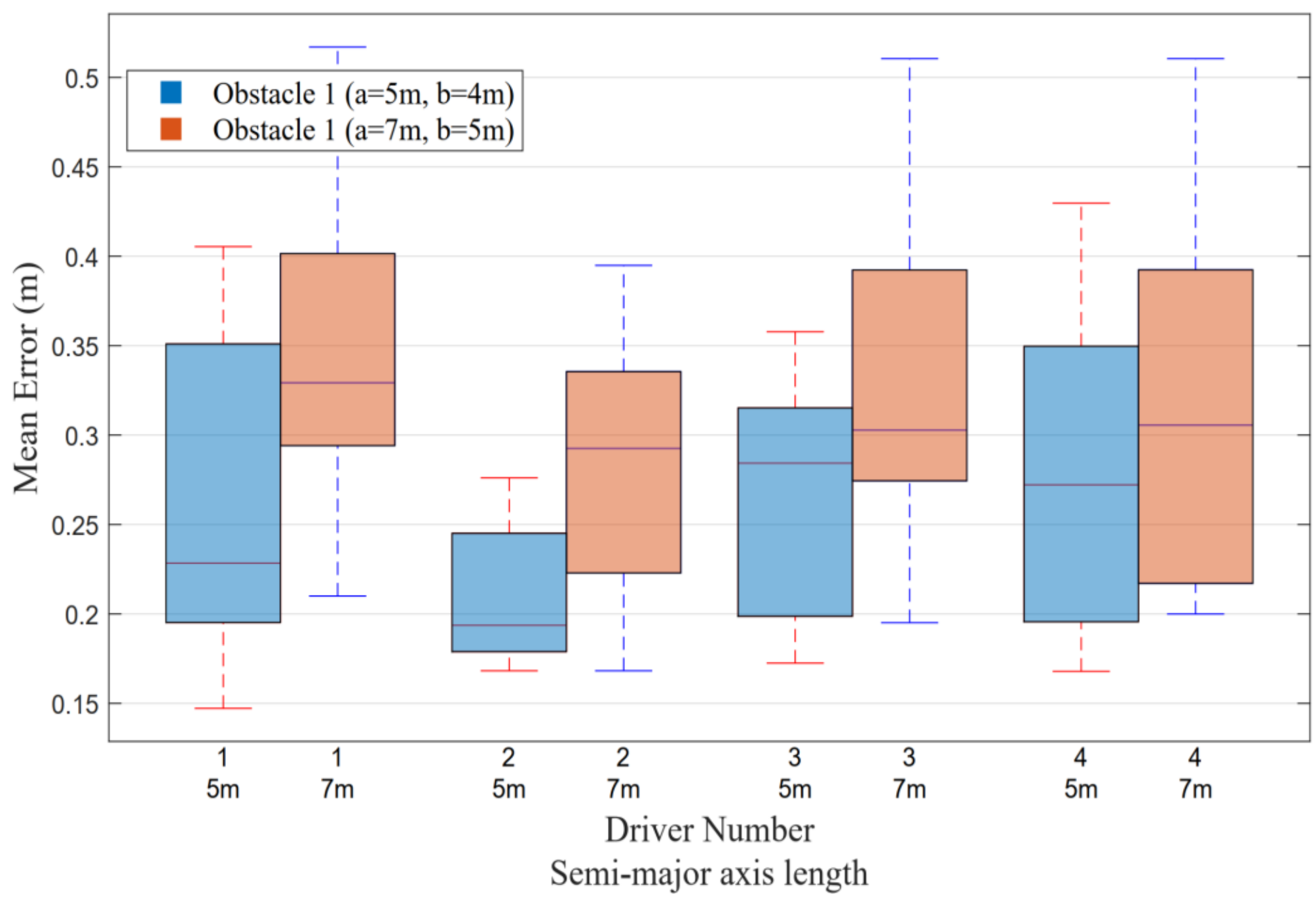

**Figure 5:** Mean trajectory error between modeled and actual trajectories for four drivers under two obstacle configurations ($a = 5\ m$, $b = 4\ m$ and $a = 7\ m$, $b = 5\ m$).

The proposed model uses the current state $\boldsymbol{x}$, collision avoidance intention $CAI$, reachability value $V_h(\boldsymbol{x})$, and obstacle envelope parameters $a, b$ as inputs, with calibrated parameters $T_m$, $T_{n1}$, $T_{n2}$, and $\alpha_{vd}$ based on driver-specific data. Figure 5 shows that the trajectory errors remain low across all drivers, with only slight increases for larger obstacles ($a = 7\ m$), likely due to higher avoidance complexity. The consistent performance across drivers and configurations highlights the model's ability to accurately capture individual driving behaviors and adapt to different obstacle scenarios.

In actual human-machine shared control scenarios, $CAI$ is estimated using the method outlined in Section III.B and does not require external initialization or updates.

***5.2 Vehicle Dynamics Model***

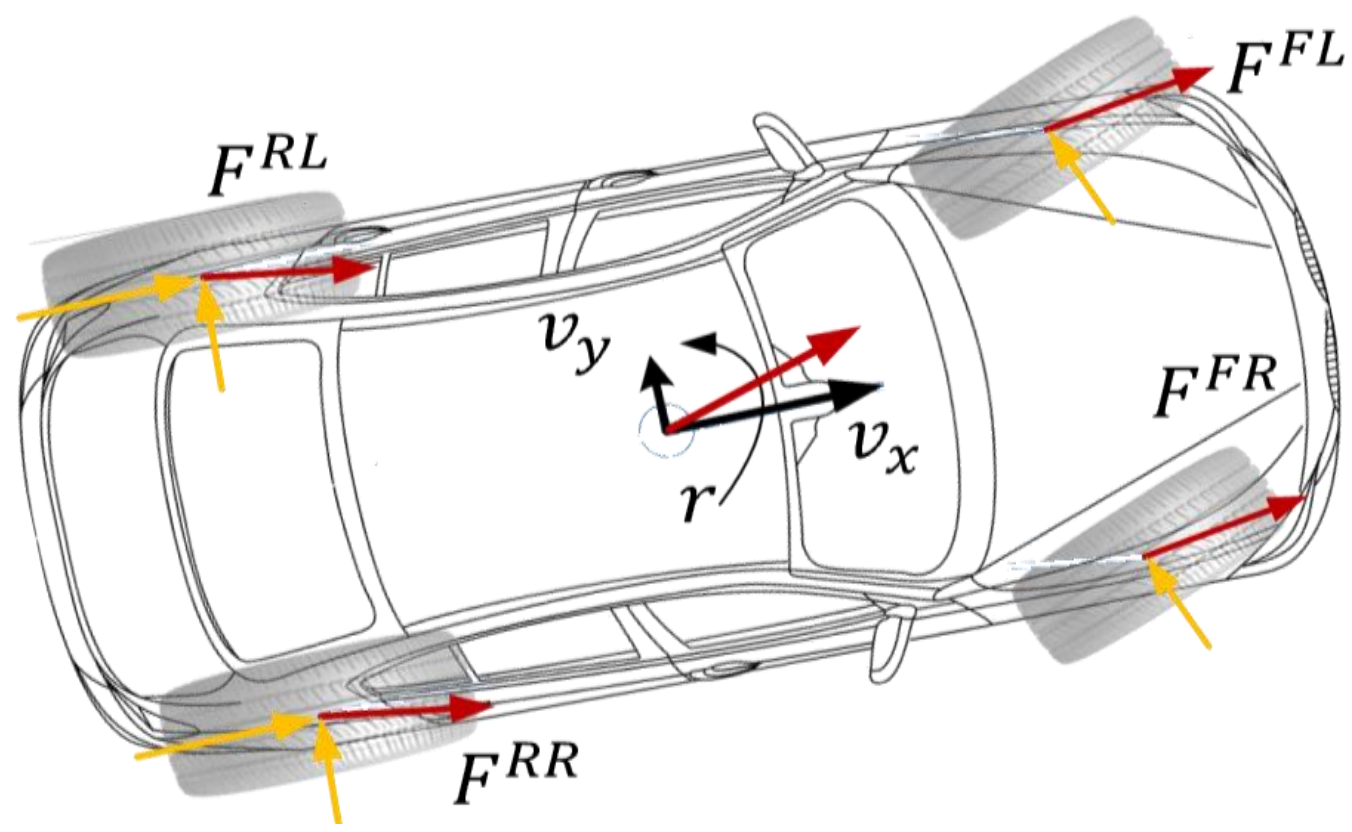

**Figure 6** Schematic representation of the vehicle dynamic model.

The purpose of the vehicle dynamics model is to provide accurate dynamic responses as input for the RL framework. The vehicle body model incorporates three primary degrees of freedom: includes longitudinal ($v_x$), lateral ($v_y$), and yaw ($r$) motions, as shown in Figure 6 [30].

$$\dot{v}_x = \frac{F_x}{m} + r \cdot v_y; \quad \dot{v}_y = -r \cdot v_x + \frac{F_y}{m} \tag{22}$$

$$\dot{r} = \frac{M_z}{I_z} \tag{23}$$

Here, $m$ refers to the total mass, $I_z$ indicates the vehicle's yaw inertia. The forces $(F_x, F_y)$ and yaw moment ($M_z$) are computed as:

$$F_x = F_x^{FL} + F_x^{FR} + F_x^{RL} + F_x^{RR} \tag{24}$$

$$F_y = F_y^{FL} + F_y^{FR} + F_y^{RL} + F_y^{RR} \tag{25}$$

$$M_z = \frac{1}{2}\left[W_f(F_x^{FL} - F_x^{FR}) + W_r(F_x^{RL} - F_x^{RR})\right] + \frac{1}{2}\left[L_a\left(F_y^{FL} - F_y^{FR}\right) + L_b\left(F_y^{RL} - F_y^{RR}\right)\right] \tag{26}$$

In these equations, $F_x^{ij}$ and $F_y^{ij}$ denote the longitudinal and lateral forces acting on each wheel,

with $i \in \{F, R\}$ for front and rear wheels, and $j \in \{L, R\}$ for left and right wheels. $W_f$, $W_r$, $L_a$, and $L_b$ are the front/rear track widths and distances from the centroid to the front/rear axles, respectively. The tire forces are computed using a modified Magic Formula model to ensure precision [30]:

$$F_x = \mu_x F_z \, sin(C_x \, tan^{-1}(B_x \phi_x)) \tag{27}$$

$$F_y = \mu_y F_z \, sin\left(C_y \, tan^{-1}\left(B_y \phi_y\right)\right) \tag{28}$$

where $\phi_x$ and $\phi_y$ are tire slip parameters defined as:

$$\phi_x = (1 - E_x)s_x + \frac{E_x}{B_x}\tan^{-1}(B_x s_x) \tag{29}$$

$$\phi_y = \left(1 - E_y\right)\delta_y + \frac{E_y}{B_y}\tan^{-1}\left(B_y \delta_y\right) \tag{30}$$

Here, $B$, $C$, $E$, and $\mu$ are parameters derived from tire data, while $s_x$ (slip ratio) and $\delta_y$ (slip angle) are calculated based on the tire input $\boldsymbol{u_t}$, which includes front-wheel steering angle $\delta_f$ and tire torque. Finally, the vehicle state is updated over a time step $\Delta T$ as:

$$\boldsymbol{x}(t + \Delta T) = \boldsymbol{x}(t) + \Delta T \cdot \dot{\boldsymbol{x}}(t) \tag{31}$$

with $\boldsymbol{x} = \left[X, Y, \varphi, v_x, v_y, r\right]$ defined in Equation 5.

***5.3 Reachability-Aware Training Framework***

The framework of Reachability-Aware Training takes in safety-critical metrics from reachability analysis into RL for adaptive machine action generation, which guides the agent to avoid collisions, reduce human-machine conflicts, and maintain control smoothness. This framework primarily involves explicit incorporation of the CARS and related metrics in the design of states, actions, and rewards.

The state space $\boldsymbol{s}$ is defined as:

$$\boldsymbol{s} = [\boldsymbol{x}, V_h(\boldsymbol{x}), CAA, CAI, \gamma, \boldsymbol{u_d}, \boldsymbol{u_m}] \tag{32}$$

At this point, $\boldsymbol{x}$ is the vehicle state as defined in Equation 5, and $V_h(\boldsymbol{x})$ is its reachability value that predicts its long-term safety of the state. The collision avoidance ability $CAA$ and intention $CAI$ offers quantitative information about environmental hazards and human input, allowing the RL agent to evaluate and complement driver behavior. The authority allocation weight, $\gamma$, reflects the human-machine control balance, and $\boldsymbol{u_d}$ and $\boldsymbol{u_m}$ are respectively the driver and machine control actions.

The action space $\boldsymbol{u_m}$ is:

$$\boldsymbol{u_m} = \delta_{mf} \tag{33}$$

where $\delta_{mf}$ represents the front-wheel steering angle. Considering the driver's experience, the controller proposed in this study refrains from intervening in braking and acceleration.

The reward function enforces safety, collaboration, and smooth control. It penalizes proximity to CARS boundaries, integrating safety prior knowledge into the RL framework. The safety reward is defined as:

$$R_{sf} = k_{sf1} \cdot exp\left(\frac{V_h(x) - k_{sf2}}{d_0}\right) \tag{34}$$

where $d_0$ is a scaling parameter representing the critical reachability threshold, and $k_{sf1}$ is a negative weighting factor emphasizing the importance of staying away from the CARS boundary. States with $V_h(x) \geq k_{sf2}$ incur significant penalties, ensuring the agent learns to avoid unsafe states.

The collaboration reward is designed to guide the agent in aligning its actions with the driver's intentions, thereby reducing human-machine conflicts. It is defined as:

$$R_{co} = -k_{co} \cdot \gamma \cdot (u_M - u_D)^2 \tag{35}$$

The penalty term $(u_M - u_D)^2$ quantifies the deviation between the machine and the driver's actions, scaled by $k_{co}$. The weight factor $\gamma$, as previously introduced, reflects the necessity of machine intervention based on the $CAA$ and $CAI$. A higher $\gamma$ value signifies limited driver ability or intention to avoid collisions, prompting the agent to prioritize safety while collaborating effectively.

The smooth control reward is designed to ensure stability and comfort by penalizing abrupt changes in machine-generated actions. It is defined as:

$$R_{sm} = -k_{sm} \cdot (u_M^t - u_M^{t-1})^2 \tag{36}$$

where $u_M^t$ and $u_M^{t-1}$ are the machine's control inputs at the current and previous time steps, respectively, and $k_{sm}$ scales the penalty. This reward promotes smooth transitions and enhances vehicle stability.

The terminal reward $R_t$ is applied at the end of an episode to evaluate the agent's overall performance, reflecting both collision avoidance and progress toward the driving goal. As stated earlier, $V_h(x) \leq 0$ is treated as a hard constraint. Violating this constraint incurs a significant penalty $k_{hc}$:

$$R_t = -k_{hc}\,,\ \ if\ \ V_h(x) > 0 \tag{37.1}$$

For episodes that conclude without a safety violation, the reward is designed to incentivize the vehicle's progress toward its driving objective:

$$R_t = k_{od} \cdot (X_t - X_r),\ \ if\ \ V_h(x) \le 0 \tag{37.2}$$

Here, $X_r$ represents the target longitudinal position the vehicle aims to reach for reference, and $X_t$ denotes the vehicle's actual longitudinal position at the end of the episode. The scaling parameter $k_{od}$ is a positive constant. If $X_t \ge X_0$ , it indicates the vehicle has surpassed the target, contributing positively to the reward. This design encourages the agent to maintain safety by avoiding states where $V_h(x) > 0$ while rewarding it for effectively reaching or exceeding the target driving goal. The overall reward for each episode can be expressed as:

$$R = R_{sf} + R_{co} + R_{sm} + R_t \tag{38}$$

***5.4 Iterative Optimization***

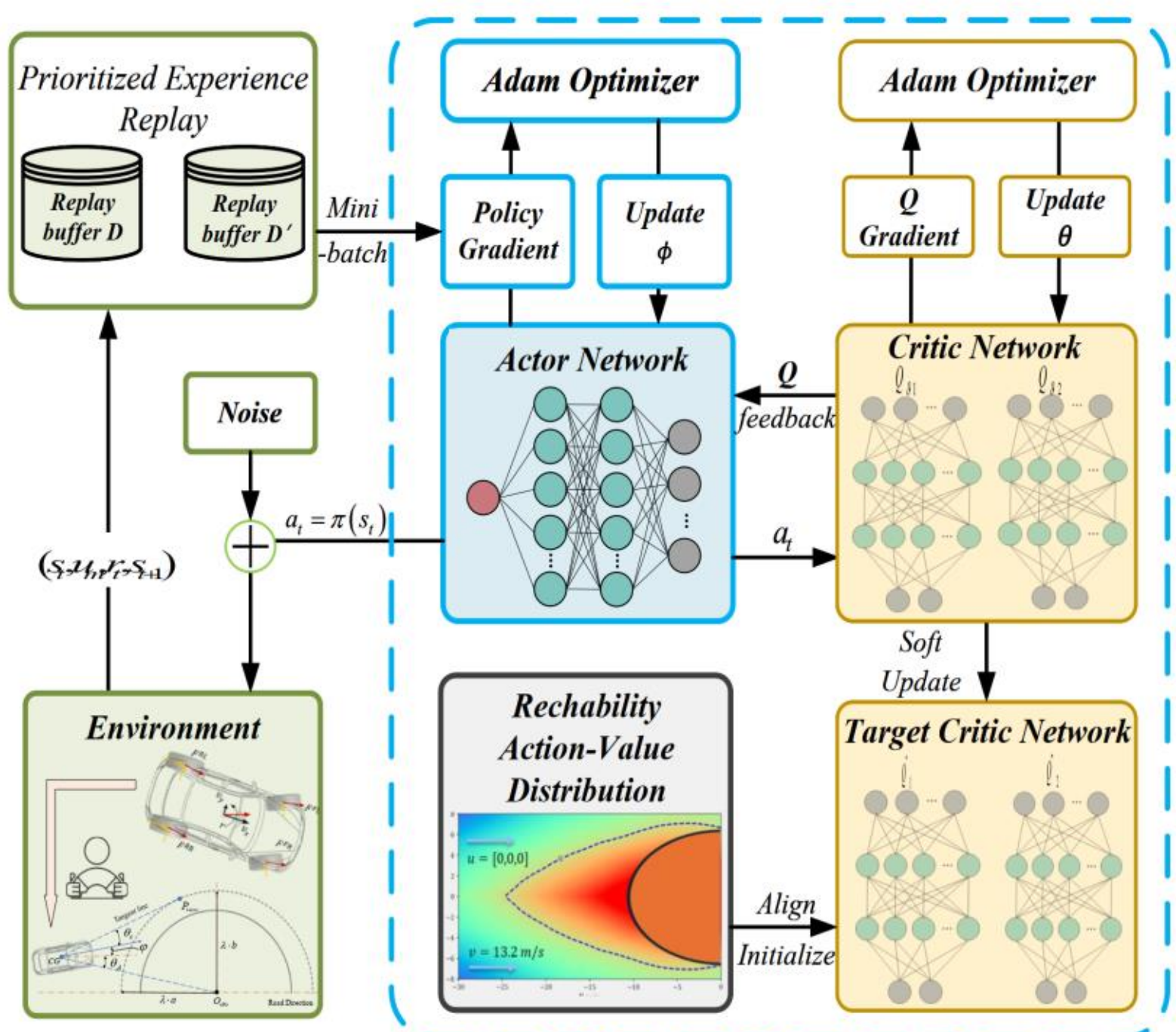

**Figure 7** Schematic diagram of the iterative optimization process.

The iterative optimization follows the Soft Actor-Critic (SAC) algorithm, combining reachability-aware initialization and safety-guided updates to enhance efficiency. As shown in Figure 7, the SAC algorithm leverages experience replay and employs double Q-functions, resulting in a total of four Q-networks (two primary Q-networks and their target networks) and one

Actor network. In our formulation, the original pair $(\boldsymbol{x}, \boldsymbol{u})$from the reachability analysis is mapped to the RL state-action pair $(\boldsymbol{s}, \boldsymbol{u_m})$. Rather than initializing the primary critic networks directly, the target Critic network $\hat{Q}$ is pre-trained on avoidance data using the reachability-based action-value function $Q_h(\boldsymbol{x}, \boldsymbol{u})$ (see Section III.A). This pre-training step incorporates collision-awareness into the target network and guides the learning process towards safer behavior from the outset. By initializing only the target network, we provide a stable and safety-aware reference for the primary Q-networks, thus stabilizing early-stage training, accelerating convergence, and improving initial performance.

The iterative optimization alternates between policy evaluation and policy improvement. During policy evaluation, the Critic parameters are updated by minimizing the soft Bellman residual:

$$L_Q = \mathbb{E}_{(\boldsymbol{s_t}, \boldsymbol{u_{mt}}, R_t, \boldsymbol{s_{t+1}})}\left[\left(Q_\vartheta(s_t, a_t) - \hat{Q}\right)^2\right] \tag{39}$$

where the target $\hat{Q}$ integrates the safety-aware reward $R$ (Equation 37) and an entropy term. This ensures effective state-action evaluation and aligns with safety priorities.

In the policy improvement step, the Actor Network $\pi_\phi$ is optimized to maximize an entropy-augmented objective:

$$L_\pi = \mathbb{E}_{\boldsymbol{s_t}, \boldsymbol{u_{mt}}}\left[\varepsilon \cdot log\, \pi_\phi(a_t \mid s_t) - Q_\vartheta(s_t, a_t)\right] \tag{40}$$

This objective balances exploitation with exploration. The temperature parameter $\varepsilon$ is adapted dynamically to keep a target entropy $\mathcal{H}_t$, preventing premature policy convergence and ensuring robust exploration.

As SAC is a widely adopted RL method, the detailed workings of the standard SAC framework are not the focus of this study and will not be elaborated upon further here.

## 6. Validation and Results

The proposed framework was validated in two critical scenarios: a sudden accident ahead and abrupt braking by the lead vehicle, both representing high-risk conditions requiring immediate collision avoidance. Driver states were categorized into three types: distracted drivers, low-attention drivers (e.g., due to intoxication), and normal drivers, reflecting varying levels of engagement and capability.

The validation process includes two stages. First, simulation results are presented, demonstrating the system's adaptability to diverse scenarios and driver collision avoidance behaviors. Next, real-world vehicle tests were conducted. These tests evaluate the framework's performance in enhancing safety and mitigating human-machine conflicts. Robustness analysis further highlights its flexibility across various driver attributes, ensuring its applicability under real-world conditions.

### *6.1 Training Results and Performance Evaluation*

During training, the Collision Avoidance Intention ($CAI$) was initialized to 0 and dynamically assigned random values within [0.2, 0.9] when $V_h(\boldsymbol{x})$ fell in the range of [-15, 0]. The obstacle was represented by an elliptical envelope with a major radius of 8 m and a minor radius of 5 m, and moved along the road at a randomly selected low speed following predefined behavioral rules. This setup emulates various levels of urgency, reflecting how drivers detect obstacles and adopt different degrees of avoidance response. Training was conducted for 6000 episodes using a system equipped with an Intel Core i9-14900HX processor and an NVIDIA 3080 GPU.

Three methods were compared. Method 1, following [31], employed Soft Actor-Critic (SAC) with vehicle states and obstacle constraints. Method 2, similar to [18], incorporated potential field-based risk distributions into the state space and treated obstacles as constraints. The proposed method integrated $V_h(\boldsymbol{x})$ into the state space and utilized the $Q_h$ to initialize the Critic network, providing safety-aware guidance during early-stage training. All methods used identical reward functions and system configurations.

Figure 8 demonstrates the comparative training performance. Our proposed method exhibited superior performance across all metrics, including convergence speed, stability, and final reward magnitude. Method 1 showed slow convergence and limited reward ceiling, suggesting difficulties in reconciling vehicle states with collision avoidance objectives. While method 2 achieved faster convergence, it reached suboptimal reward levels. The slightly higher machine intervention levels in Method 2 may have negatively impacted original task performance. The proposed method achieved optimal rewards and smoother training progression.

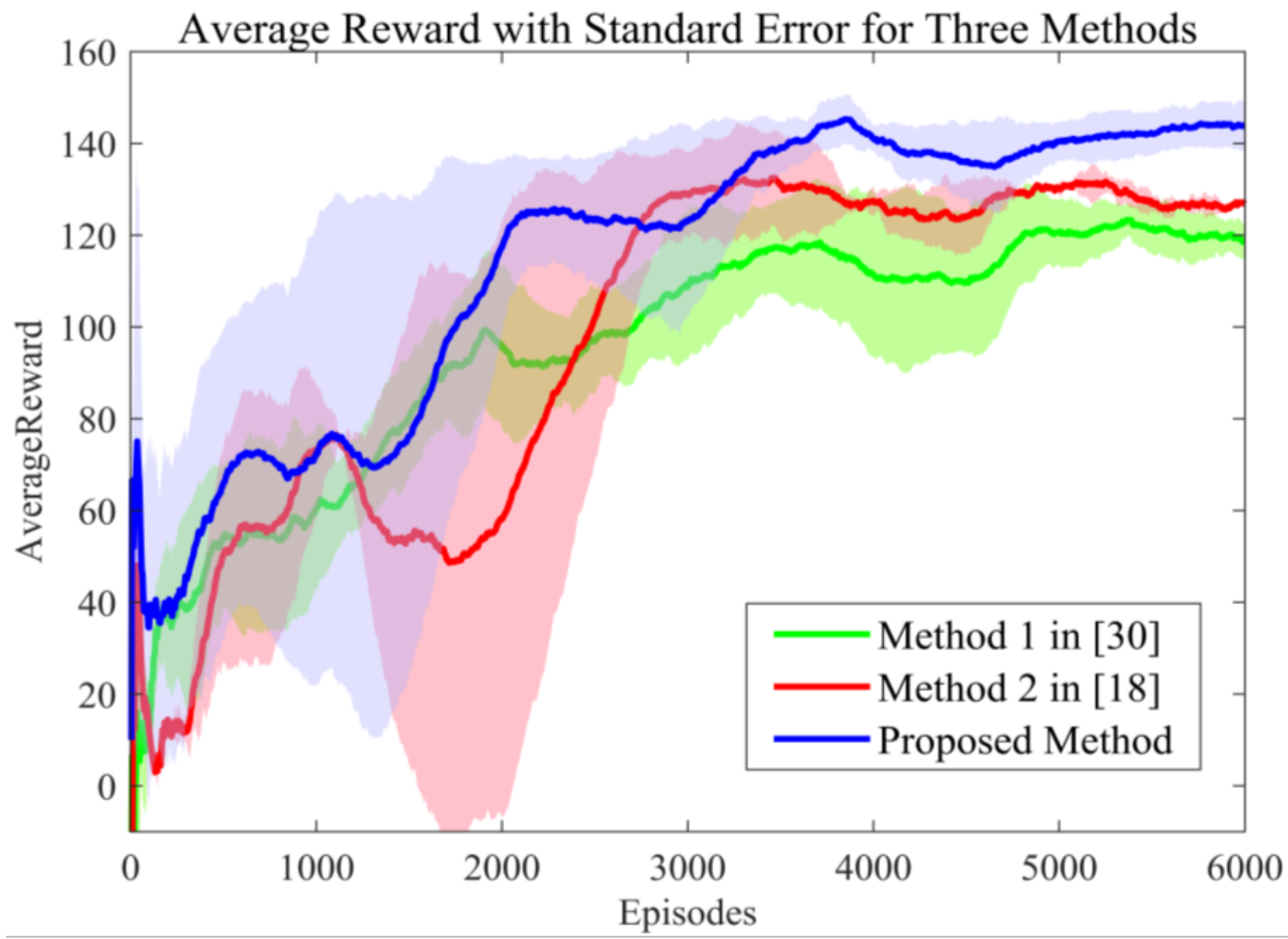

**Figure 8** Average reward with standard error for three methods.

### *6.2 Simulation Experiments with Dynamic Obstacles*

In the simulation experiments, we selected a one-way road high-risk driving scenario. The dynamic obstacle, modeled as the elliptical envelope described in the previous section, is initially positioned 15 meters ahead and 6 meters to the right of the ego vehicle. It moves forward at an initial speed of 10 m/s and subsequently decelerates at 6 m/s² while performing a leftward lane change to avoid a potential hazard. This setup emulates a realistic dynamic collision risk where the obstacle trajectory evolves unpredictably.

Three control strategies were evaluated: the rule-based AEB baseline that applies full braking below a constant threshold, the learnable baseline in [18], and the proposed reachability-aware shared-control policy. Driver cooperation was parameterised by the Collision-Avoidance Intention (CAI). We first fixed CAI = 0.3 to emulate a distracted driver for whom stand-alone human control is almost certain to fail, then repeated the experiment with CAI = 0.7 to quantify human–machine conflict when the driver is fully engaged. Performance metrics were (i) success rate, (ii) Maximum Forward Distance (MFD) along the prescribed route after the evasive manoeuvre, and (iii) Mean Normalised Steering-Angle Difference (MNSAD) between human and machine commands.

Table I summarises the quantitative results for the triggering threshold $V_h(x) = -10$. Under low CAI the proposed method achieved the highest success rate (93.4 %) while maintaining the longest

MFD (77.8 m) and a MNSAD comparable to the baselines, demonstrating its ability to compensate for absent driver input without unduly sacrificing driving objectives. When CAI = 0.7 all methods succeeded, yet the proposed policy achieves the best original driving task performance (MFD = 109.5 m) and intervened the smallest steering correction from the driver (MNSAD = 17.9 %), confirming that it intervenes only when strictly necessary and otherwise yields authority to the human.

**Table I** Performance Comparison of Collision Avoidance Methods Under Different *CAI* Levels

| CAI | Method | Success rate | M.F.D. | M.N.S.A.D. |
|---|---|---|---|---|
| 0.3 | Method 1 [31] | 83.7% | 70.9m | 71.8% |
| | Method 2 [18] | 89.2% | 74.9m | 74.9% |
| | Our Method | 93.4% | 77.8m | 74.6% |
| 0.7 | Method 1 [31] | 100% | 106.5m | 24.1% |
| | Method 2 [18] | 100% | 108.4m | 26.3% |
| | Our Method | 100% | 109.5m | 17.9% |

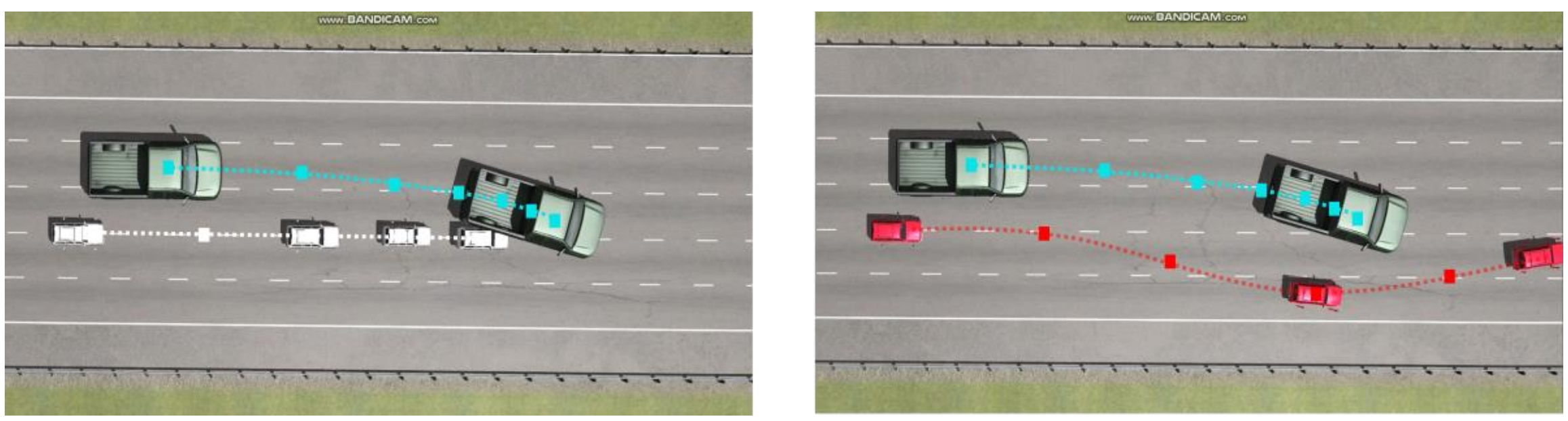

**Figure 9.** Collision-avoidance trajectories in the dynamic-obstacle scenario.

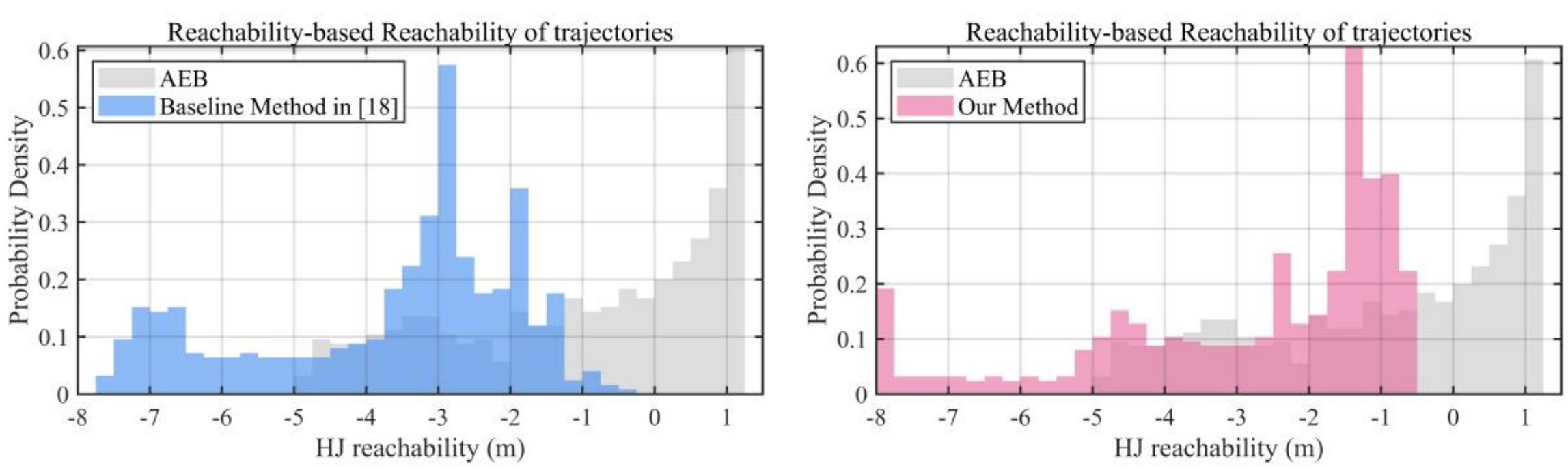

**Figure 10.** Probability-density histograms of Hamilton–Jacobi reachability $V_h$ evaluated along the avoidance trajectories.

Figure 9 depicts representative trajectories produced by the reachability-aware controller (red) and the AEB baseline (grey). The proposed controller initiates a gentle lateral detour before deceleration, preserving forward progress and maintaining a noticeably clearance to the obstacle. In contrast, AEB performs an abrupt straight-line stop, bringing the vehicle closer to the danger zone and ultimately causing a collision.

Figure 10 reveals the risk footprint during the collision avoidance process. A probability density analysis is conducted over the reachability values V(x) along the avoidance trajectories. The AEB method shows a large portion of states distributed in the danger zone, indicating actual collisions. The baseline method [18] concentrates its reachability values around –3.0 m, while the proposed method shifts the peak density toward –1.5 m. Moreover, compared to baseline method, our method exhibits a lower peak risk. As a result, most states under our method fall within a moderate to low safety margin, suggesting that our approach achieves safety with only the necessary level of intervention.

Altogether, embedding the reachability index and CARS constraint into both the state and reward furnishes the agent with a physically meaningful safety signal. The resulting policy augments human drivers when they under-react, respects them when they over-perform, and strikes a tight balance between risk reduction and task preservation.

### *6.3 Real-World Experiments with Sudden Static Obstacles*

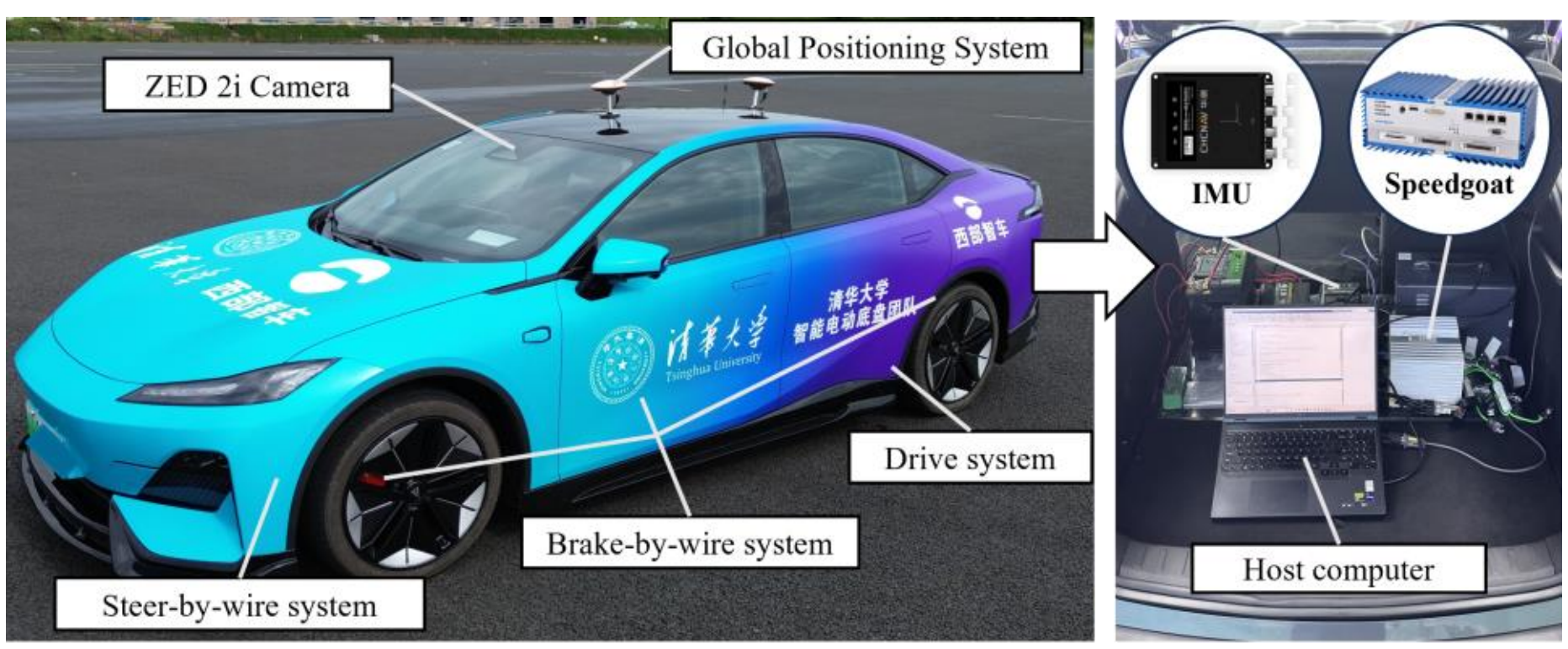

**Figure 11.** Test vehicle platform and its key components

For real-world testing, the vehicle platform is an actual-size commercial-grade electric vehicle, as illustrated in Figure 11 and Table II. It features steer-by-wire electric power steering to allow closed-loop control of the steering from the front wheels; the drive system performs closed-loop control of torque; its hydraulic brake system has brake-by-wire for controlling master cylinder pressure in a closed loop. The platform is further equipped with inertial navigation (CGI 610), GNSS, and onboard cameras to provide real-time measurements of vehicle position, velocities, accelerations, yaw rate, and wheel speeds.

All sensor signals are transmitted via a CAN FD network to a Speedgoat real-time industrial computer, which serves as the central controller. The Speedgoat system is equipped with an Intel i7 2.5 GHz dual-core CPU, 8 GB of RAM, and high-speed I/O modules, providing sufficient computational capacity to run the trained reinforcement learning policy in real time. The control algorithm is developed using MATLAB/Simulink and automatically compiled for deployment onto the Speedgoat. During operation, the Speedgoat receives vehicle state data and computes control commands (including steering angle, torque request, and brake pressure), which are then transmitted back to the vehicle's subsystems via CAN FD.

This closed-loop system enables fully onboard, real-time execution of the reinforcement learning–based shared control framework with a control cycle of 25 ms, without relying on any external computing or cloud-edge collaboration.

We conducted experiments on a closed road to simulate an emergency scenario where a traffic accident suddenly occurs ahead under low-visibility conditions, such as heavy fog. A static obstacle, representing the crashed vehicle, appears unexpectedly in the driving lane and is detected only when its edge is 15 meters away from the ego vehicle. The obstacle is modeled as an elliptical envelope with a semi-major axis of 8 meters and a semi-minor axis of 6 meters, approximating the occupied road area. The driver's response in this scenario is characterized by insufficient steering intent to avoid the collision, and the corresponding control inputs are provided by the recorded action sequence of a real human driver. For comparison, we also tested a baseline method like [18], which used the obstacle region as a constraint, incorporated the risk from the artificial potential field into the state, and replaced the safety reward design in Formula 33. To ensure fairness, the driver inputs in the baseline method were directly obtained through a lookup table matching the driver inputs in our method. The original target longitudinal speed during the trials was 12.5m/s.

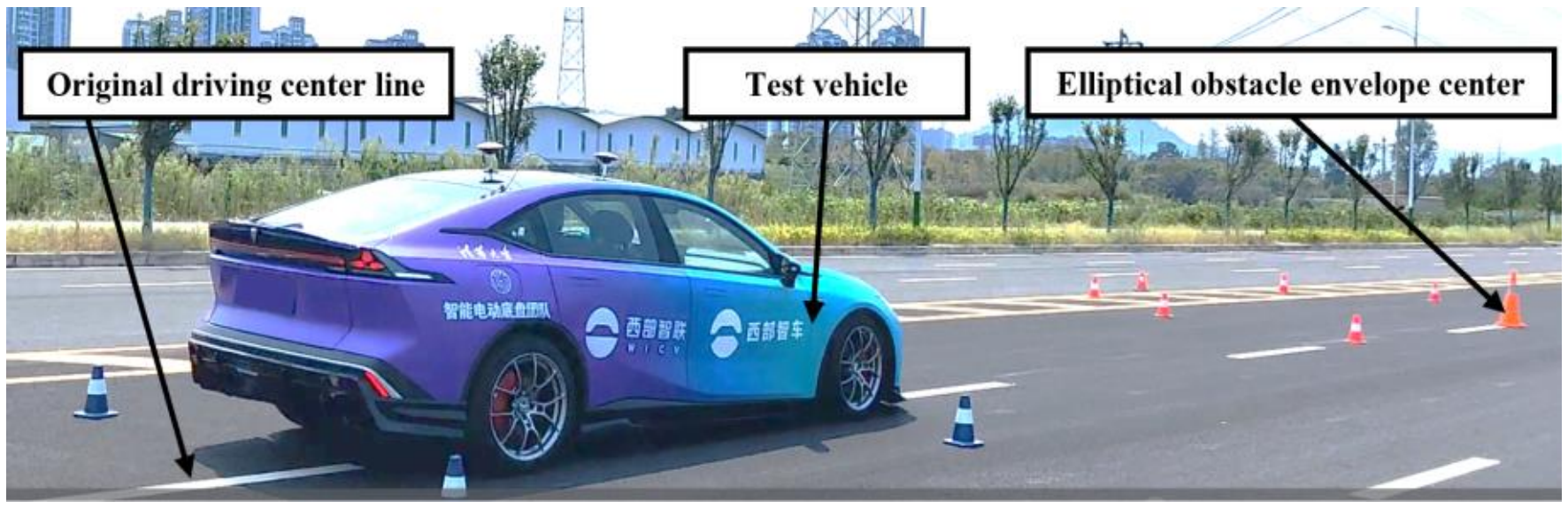

**Figure 12.** Scene settings in real vehicle testing

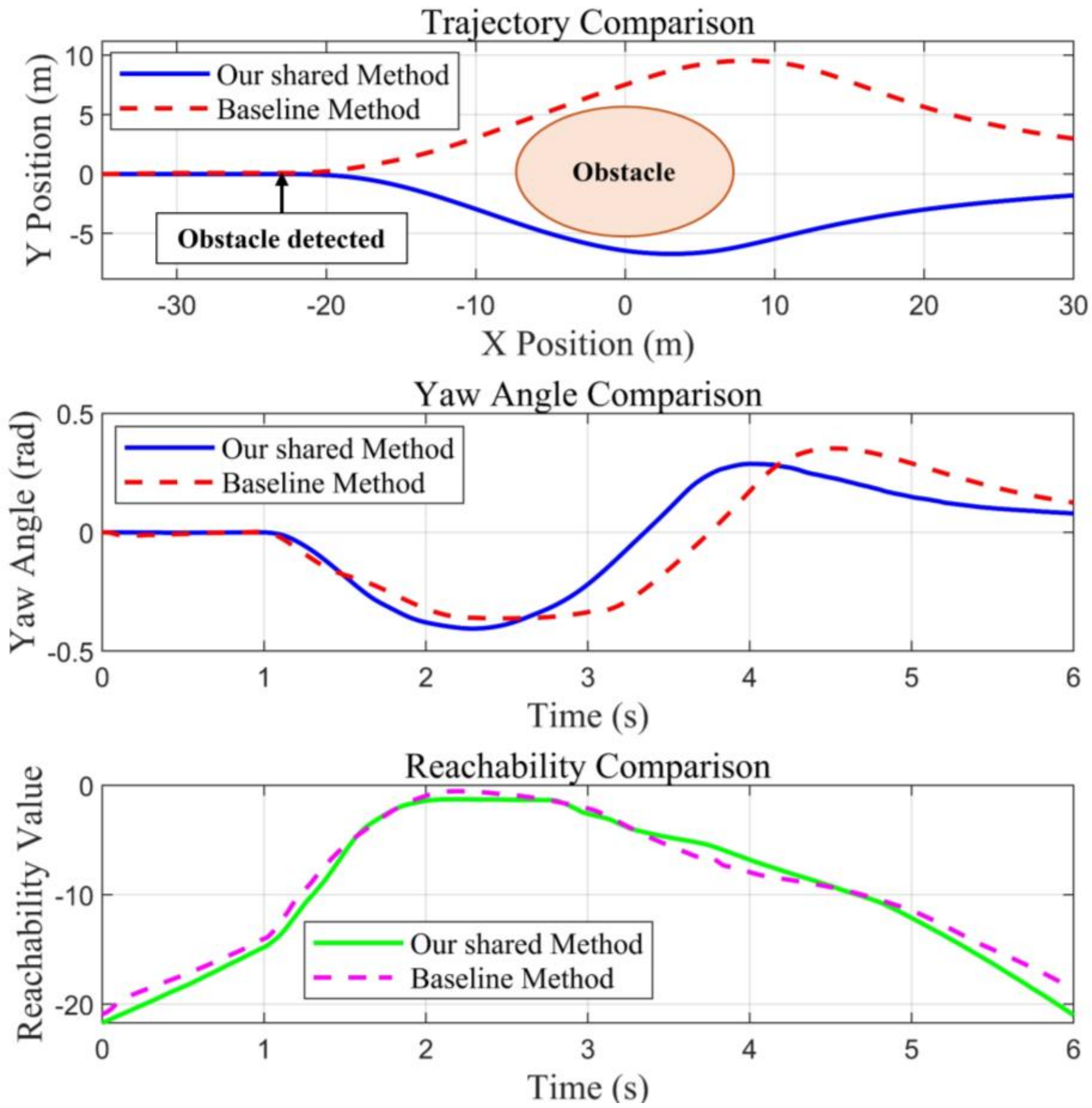

**Figure 13.** Comparison of vehicle status in the collision avoidance test

**Table II** The Employed Vehicle Parameters

| Parameter | Value | Parameter | Value |
|---|---|---|---|
| Total Mass ($kg$) | 1615 | Initial Speed ($m/s$) | 12.5 |
| Roll inertia ($kg.m^2$) | 251 | Friction coefficient | 0.75-0.9 |
| Yaw inertia ($kg.m^2$) | 2795 | Tire | 215/55-R17 |

To evaluate the performance of the proposed human-machine shared control system, the experimental results were analyzed with a focus on the vehicle's trajectory, yaw angle, and control inputs. The first sub-figure of Figure 13 lists the trajectory diagrams during the obstacle avoidance process. The proposed human-machine shared control is less disturbed by obstacles and travels 4.5m more in the same 6s time compared to baseline driving. This can be explained by the smaller lateral deviation of shared control, while the trajectory of the vehicle controlled by humans alone has a more obvious lateral deviation, which may be caused by the prolonged reaction time and the inability to recognize the dangerous changes of obstacles in time. The comparison of yaw angles in the second sub-figure further supports the above results. Under the proposed shared control, the yaw angle adjustment is smoother, avoiding the overshooting seen with human control.

The third sub-figure of Figure 13 shows the change of reachability value during obstacle avoidance. In the initial stage, the reachability values of the two methods are similar, indicating that the safety level is comparable. When approaching the obstacle (about t=1 second), the reachability value of the baseline method rises sharply, indicating a delayed response to the collision risk. In contrast, the reachability value of our shared control method rises steadily and reaches a lower peak value (about t=2 seconds). This is due to the integration of reachability heuristics and CARS hard constraints, which enables the system to predict theoretical collision risks and effectively adjust operation, achieving safer trajectory.

Figure 14 illustrates the changes in control variables during the human-machine co-control process, including the final steering input, driver input, machine control input, and control weights. In the first subplot, the green line in the first sub-figure is the final control input, which reflects an overall smooth steering process. In the second subplot, the blue dashed line indicates the driver input, while the red line shows the machine intervention input. Although some oscillations in the machine intervention input are observed before the collision avoidance begins and near its completion, the control weights reduce its impact on the final input. The analysis reveals that at critical moments of collision avoidance, our controller intervenes rapidly and significantly, compensating for the driver’s insufficient reaction to obstacles. During the mid-stage of collision avoidance, the controller reduces the steering magnitude while ensuring safety and assists the vehicle in realigning promptly. This helps to reduce the lateral displacement amplitude while ensuring collision avoidance, avoid other potential dangers and instability, and promote the performance of the original straight-line driving performance.

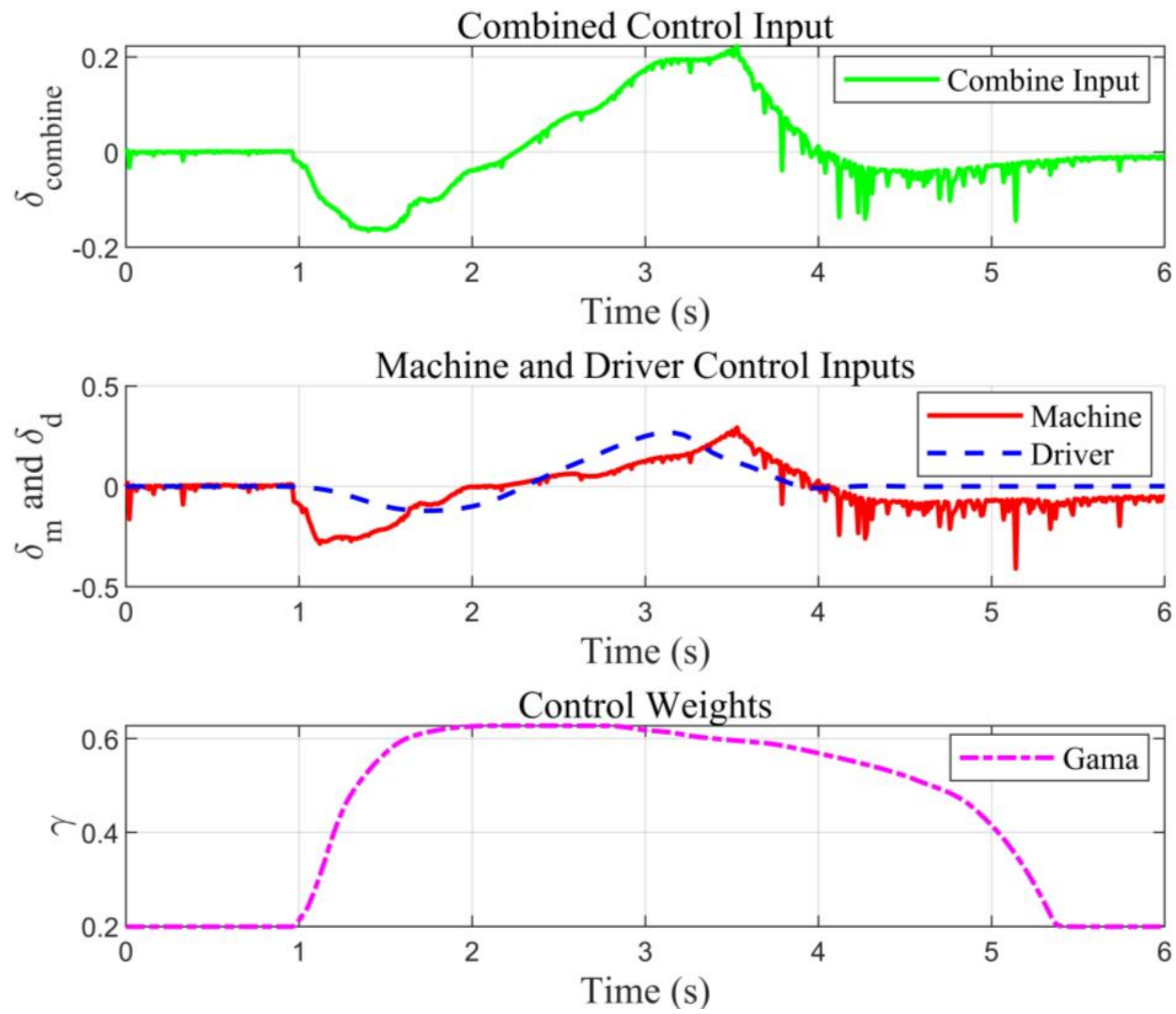

**Figure 14.** Control input and control weights in collision avoidance test

In addition to the above tests, we also conducted tests under a more critical emergency collision avoidance scenario where the collision avoidance was activated at the instant when the value function $V_h(\boldsymbol{x})$ reached -1, indicating that a crash was nearly inevitable. In this scenario, both the driver and the controller performed coordinated steering and braking maneuvers to avert severe impact. Due to space constraints, results for both our method and the baseline approach in these high-risk scenarios are provided as Table III and **supplementary video material**. Remarkably, our proposed approach successfully prevented any collision, whereas the baseline method, under suboptimal driver input conditions, resulted in a slight side-scrape. Under more extreme collision risk, the benefits delivered by CARS hard constraints became even more pronounced.

To assess real-time performance, we analyzed the Speedgoat system logs during deployment. The controller runs at a fixed 25 ms cycle, with an average computation time of 4.91 ms, minimum 2.60 ms, and maximum 8.79 ms. The average CPU load was 39%, indicating stable execution with sufficient computational headroom. This confirms that the RL-based shared control framework can be executed fully onboard without external computing.

**Table III** Results Under High-Risk Emergency Collision Avoidance Scenario (Triggered at $V_h(\boldsymbol{x}) = -1$)

| Driver | Approach | Collision Outcome | M.N.S.A.D. |
|---|---|---|---|
| 1 | Baseline | Slight Side-Scrape | 51.8% |
| | Proposed | No Collision | 54.6% |
| 2 | Baseline | No Collision | 48.4% |
| | Proposed | No Collision | 43.9% |

### *6.4 Robustness to Variations in Driver Characteristics*

To further evaluate the adaptability and reliability of the proposed human-machine shared control framework, we now focus on its robustness to changes in driver behavior. Building upon the driver model introduced in Section IV.A, we conduct a simulation analysis to examine how parameter variations—particularly in the driver's b rain reaction time $T_m$ and key parameter of preview angle $k_\lambda$—influence the controller's performance.

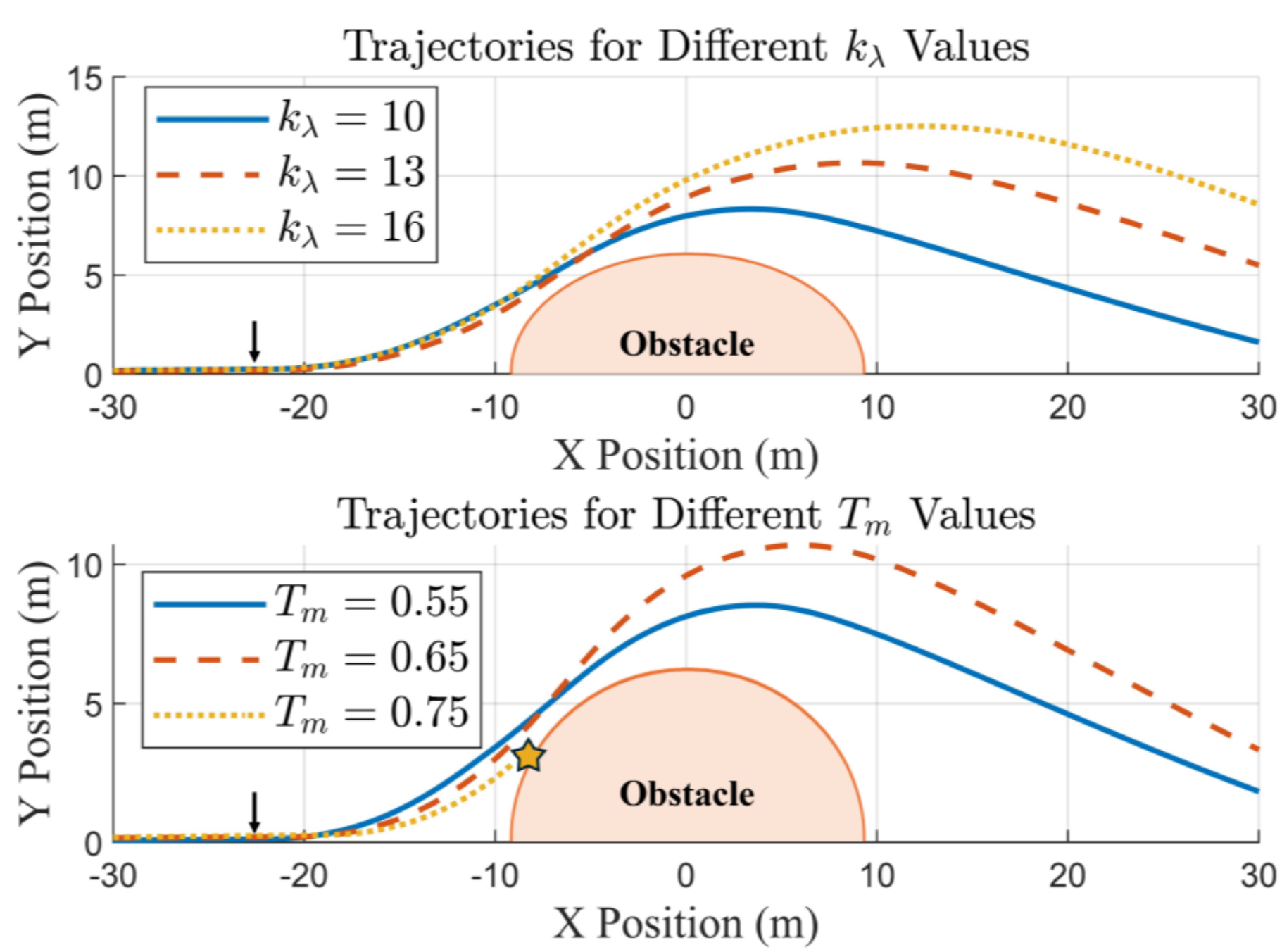

**Figure 15.** Influence of driver preview parameter $k_\lambda$ and response time $T_m$ on vehicle trajectories.

Figure 15 shows the impact of changes in driver characteristics $k_\lambda$ and $T_m$ on the human-machine shared collision avoidance trajectory. The results indicate that the driver's preview characteristic $k_\lambda$ primarily influences the trade-off in the avoidance strategy—balancing adherence to the original route against the need to circumvent potential hazards. Despite these variations, our approach consistently ensures that the vehicle state does not enter infeasible regions.

In contrast, the driver's response time $T_m$ exerts a more direct impact on collision avoidance outcomes. For instance, when $T_m$ reaches 0.75s, the vehicle initiates emergency braking to mitigate collision risks, abandoning the original driving objective. This highlights that while preview parameters shape the strategic balance between path fidelity and safety, prolonged response delays prompt the system to prioritize immediate risk reduction over maintaining the initial task.

## 7. Discussion

This study aims directly at enhancing accident prevention and road safety through a reachability-aware, data-driven reinforcement learning framework. By approximating CARS using large-scale data and RL, we enabled proactive identification of high-risk states and embedded CARS safety constraints that prevent the system from entering collision-prone regions. This reachability-inspired mechanism acts as a predictive safety barrier, providing robust protection even under low-engagement driving conditions.

We leverage reachability analysis to evaluate both the vehicle's collision avoidance ability and driver's collision avoidance intention, and design an interpretable control authority allocation mechanism accordingly. This allows the system to intervene only when necessary, ensuring collision avoidance while minimizing human-machine conflict. Experimental results in high-risk scenarios demonstrate that the proposed framework can effectively prevent accidents without compromising task performance, showing strong potential for real-world deployment in advanced driver-assistance systems (ADAS).

Future work will focus on enhancing the framework's adaptability to a wider range of driver behaviors and dynamically evolving traffic scenarios, with the goal of further improving accident prevention under real-world uncertainties.

## Acknowledgement

This work was supported by the National Key R&D Program of China under Grant No.2022YFB2503103.

## Appendix

Accompanying this paper is a supplementary video showcasing the real vehicle tests conducted in this study, which includes scenarios on both normal and emergent conditions.